\documentclass[10pt,twocolumn,letterpaper]{article}

\usepackage{wacv}              % To produce the CAMERA-READY version
\usepackage{multirow}
\definecolor{wacvblue}{rgb}{0.21,0.49,0.74}
\usepackage[pagebackref,breaklinks,colorlinks,allcolors=wacvblue]{hyperref}

\def\wacvPaperID{1348} % *** Enter the WACV Paper ID here
\def\confName{WACV}
\def\confYear{2026}

\title{Intraoperative 2D/3D Registration via Spherical Similarity Learning and Differentiable Levenberg-Marquardt Optimization}

\author{Minheng Chen\textsuperscript{12}\footnotemark[1],~~~Youyong Kong\textsuperscript{1}\footnotemark[2]\\
\textsuperscript{1} Southeast University
\textsuperscript{2} University of Texas at Arlington\\
\tt\small {\{mh$\_$chen, kongyouyong\}@seu.edu.cn}\\
}

\begin{document}
\maketitle
\renewcommand{\thefootnote}{\fnsymbol{footnote}}
\footnotetext[1]{Work done during undergraduate studies at Southeast University.}
\renewcommand{\thefootnote}{\arabic{footnote}}
\renewcommand{\thefootnote}{\fnsymbol{footnote}}
\footnotetext[2]{Corresponding author.}
\renewcommand{\thefootnote}{\arabic{footnote}}
\begin{abstract}
% Intraoperative 2D/3D registration aligns preoperative 3D volumes with real-time 2D radiographs, enabling accurate overlay of additional auxiliary anatomical information that is not visible in intraoperative imaging onto the surgical scene. This provides precise localization of instruments and implants, enhancing surgical accuracy and safety.
% A recently proposed fully differentiable similarity learning framework, which enables neural networks to approximate the geodesic distance between two poses on the manifold in SE(3), has garnered considerable attention. 
% It greatly increases the capture range of registration and mitigates the effects of substantial disturbances on registration.
Intraoperative 2D/3D registration aligns preoperative 3D volumes with real-time 2D radiographs, enabling accurate localization of instruments and implants.
A recent fully differentiable similarity learning framework approximates geodesic distances on SE(3), expanding the capture range of registration and mitigating the effects of substantial disturbances, but existing Euclidean approximations distort manifold structure and slow convergence.
% However, existing methods approximate manifold in Riemannian geometry within Euclidean space, leading to inaccurate portrayal of manifold's local structure, with a lengthy convergence process. 
To address the above limitations, we explore similarity learning on non-Euclidean spherical feature spaces to improve the ability to capture and fit complex manifold features.
We extract feature embeddings using a CNN-Transformer encoder, project them into spherical space, and approximate their geodesic distances with Riemannian geodesic distances in the bi-invariant SO(4) space. This enables the learning of a more expressive and geometrically consistent deep similarity metric, enhancing the network’s ability to distinguish subtle pose differences.
Fully differentiable Levenberg-Marquardt optimization is adopted to replace the existing gradient descent method to accelerate the convergence of the search during inference phase.
% Extensive experiments and ablation studies on real and synthetic datasets demonstrate that our approach achieves superior registration accuracy in both patient-specific and patient-agnostic scenarios. 
Experiments on real and synthetic datasets show superior accuracy in both patient-specific and patient-agnostic scenarios.
\end{abstract}
 
\begin{figure*}[h!]
\centering
\includegraphics[width=\linewidth]{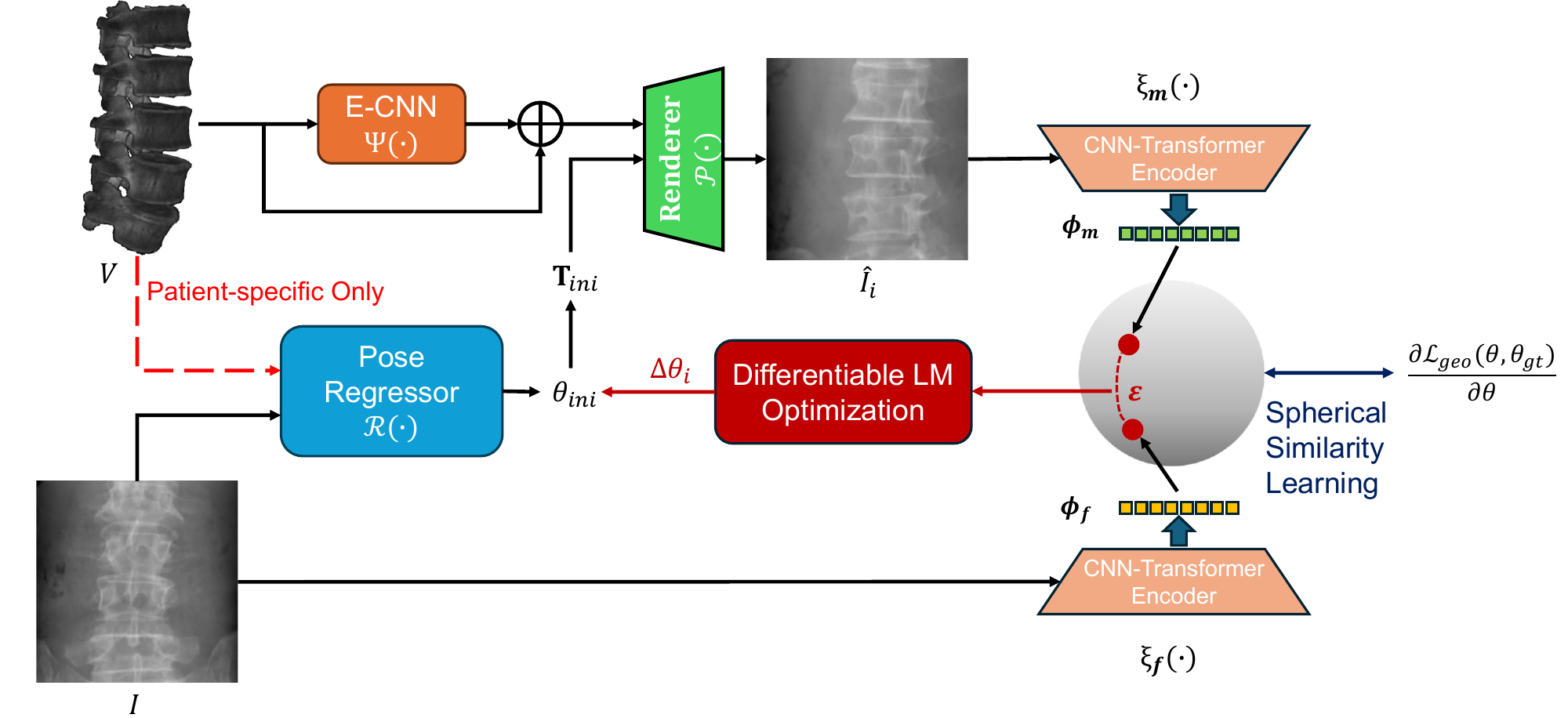}
\caption{
    \textbf{Overview of the proposed framework.}
    We first employ a regressor to initialize the pose and then refine it using differentiable Levenberg-Marquardt optimization based on spherical similarity learning. Spherical similarity learning consists of two main components: extracting image feature representations using CNN-Transformer encoders $\xi(\cdot)$ and projecting these embeddings into hypersphere space, where the geodesic distance between them is computed as a measure of deep similarity. During training, we enforce the gradient of this deep similarity with respect to $\theta$ to approximate the gradient of the geodesic distance between $\theta$ and the ground truth $\theta_{gt}$ in SE(3).
}
\label{fig1}
\end{figure*}
\section{Introduction}
\label{sec:intro}
Intraoperative 2D/3D registration is a process aimed at aligning intraoperative 2D images, such as X-ray images, with corresponding preoperative CT scans. It is a crucial step in providing surgical planning guidance and navigation positioning for minimally invasive procedures. This process allows for the precise placement of implants or surgical instruments during spinal surgeries, such as percutaneous vertebroplasty and pedicle screw internal fixation~\cite{unberath2021impact}.

% Because of the modal differences between CT and X-ray data, it is usually necessary to project the 3D scan to obtain a 2D image, or to reconstruct a 3D volume from the 2D image to unify the dimensions between the images to be registered.
In conventional optimization-based registration methods~\cite{grupp2018patch,penney1998comparison,knaan2003effective,frysch2021novel,de20163d,markelj2012review}, diverse ray-tracing techniques are employed to produce simulated two-dimensional X-ray images, commonly referred to as digitally reconstructed radiographs (DRRs), from the 3D CT volume. These techniques simulate the attenuation of X-rays within the human body, providing a crucial component for accurate registration. 
The similarity between DRRs and X-rays is then evaluated by using some statistical-based similarity measures, \textit{ie.}, normalized cross-correlation (NCC) and mutual information (MI) or local feature representations~\cite{toews2017phantomless}. Derivative-free optimization techniques, such as BOBYQA~\cite{powell2009bobyqa} and CMA-ES~\cite{hansen2001completely}, are employed to explore the similarity function and identify both the minimum value and the corresponding patient pose within the solution space.
Yet, such similarity functions are not globally convex in the search space, consequently presenting numerous local minima.
Therefore, it is frequently employed alongside complex multi-resolution optimization strategies or necessitates a robust initialization.

In literature, with the development of machine learning technology,  several regression-based~\cite{li2025automatic,miao2016cnn,leroy2023structuregnet,zhao2024automatic} and landmark-based 
methods~\cite{esteban2019towards,grimm2021pose,brandstatter2024rigid,markova2022global} have been proposed, demonstrating promising performance. However, they suffer from either the vulnerability to minor image changes, leading to prediction instability, or demand an extensive amount of clinically unsatisfactory and nearly impractical well-annotated data for training. 
These limitations significantly hamper the clinical applicability of these methods.
Recently, inspired by the traditional optimization-based methods, Gao \textit{et al.}~\cite{gao2020generalizing,gao2023fully} proposed a fully differentiable framework that approximates the geodesic between two points in a special Euclidean group in three dimensions (SE(3)) by using neural networks. 
This framework increases the capture range of registration while being more resilient to changes in local features. 
This constrained similarity learning approach greatly mitigates disturbances caused by large offsets and image noise, demonstrating robustness. 
Besides, an initial study~\cite{chen2024fully} attempted to use a correlation-driven method to decompose local and global features during the similarity learning process.
However,  existing similarity learning-based frameworks still have several shortcomings. We observed that current methods opt to approximate manifold in Riemannian geometry within Euclidean space. However, Riemannian manifolds exhibit intricate local geometric structures. Approximating using Euclidean space leads to an inaccurate portrayal of the manifold's local structure~\cite{zhao2023spherical}. 
Because distances in feature space ought to be calculated along manifold directions, and the Euclidean distance serves as merely a suitable approximation for small distances.
And we could not determine the precise distance metric in the latent space that is best suited for approximating geodesics on the special Euclidean group.
Moreover, the convergence process of existing methods tends to be relatively protracted (see results in~\cref{tab::non-patient-specific}).

In this work, we propose a more reasonable paradigm for the similarity learning-based 2D/3D registration. First, to mitigate the inaccuracies in local geometric descriptions when approximating Riemannian geodesic distance in Euclidean space, we explore alternative metric approximations in non-Euclidean embedding spaces and extend them to special orthogonal groups with bi-invariance. This means that the distance between two elements remains unchanged under both left and right multiplications, whereas SE(3) is only left-invariant. 
As a result, the landscape of the learned similarity metric becomes unambiguous, more contrastive, and smoother, enabling a broader capture range while preventing premature convergence.
During experiments, we found that the similarity measure acquired through similarity learning in hyperspherical space yields the most effective registration outcomes, rectifying the inherent ambiguity of single-view 2D/3D registration.
We introduce a differentiable Levenberg-Marquardt (LM) optimization strategy as an alternative to the gradient descent method to accelerate the convergence of registration. 
% To the best of our knowledge, this is the \textit{first} time that the differentiable LM algorithm has been applied in CT/X-ray registration.
By conducting studies on pelvic data, spine data, and clinical data on endovascular thrombectomy for acute ischemic stroke,
we demonstrate the superiority of our approach in the context of intraoperative CT to X-ray registration. 
Our method achieves state-of-the-art performance across multiple datasets, including pelvic and spine datasets, as well as clinical data for cerebral endovascular therapy, demonstrating its robustness and effectiveness in diverse 2D/3D registration scenarios.
It should be noted that, compared with single-view registration, multi-view or biplanar 2D fluoroscopic registration to a 3D CT volume provides better global convexity and benefits from well-established solvers~\cite{mi2023sgreg,zollei20012d,liao2019multiview}. However, it requires acquiring additional fluoroscopic images, leading to higher radiation exposure that may harm patient health. In this work, we focus on single-view 2D/3D registration, aiming to balance accuracy with clinical safety.

\section{Related Work}
% \paragraph{Optimization-based 2D/3D registration.}
\subsection{Learning-based 2D/3D Registration}
Learning-based methods in 2D/3D registration are primarily categorized into regression-based and landmark-based approaches. Regression-based methods~\cite{guo2021end,leroy2023structuregnet,miao2016real,zhao2024automatic}, directly estimate human pose from 3D CT volumes or paired 2D X-ray/DRR images. The primary advantage of this approach lies in its ability to achieve real-time posture estimation, making it particularly suitable for time-sensitive scenarios, \textit{e.g.} intraoperative navigation and robotic-assisted surgery. However, its major limitation is relatively lower accuracy compared to landmark-based methods, which may impact precision in high-stakes medical scenarios requiring meticulous anatomical alignment.
The landmark-based approach~\cite{esteban2019towards,grimm2021pose,liao2019multiview,shrestha2023x,shrestha2024rayemb} formulates this task as a well-established perspective-n-point (PnP) problem~\cite{li2012robust}, which determines the patient's actual posture by estimating the spatial transformation of corresponding landmarks between the reference and moving images.
Additionally, a point-to-plane correspondence solver~\cite{jaganathan2023self,schaffert2020learning,wang2017dynamic,jaganathan2021deep} has been proposed as an alternative approach to address this problem. While these methods improve alignment accuracy, it is highly dependent on the precision and visibility of landmarks~\cite{toews2017phantomless}, making them susceptible to noise, occlusions, and variations in image resolution. Furthermore, to enhance robustness, sophisticated techniques such as RANSAC~\cite{fischler1981random} are often required, leading to increased computational cost and processing time.
Recently, some studies~\cite{chen2024embedded,zhang2023patient,gopalakrishnan2024intraoperative,downs2025improving,gopalakrishnan2025rapid} have adopted a two-stage approach to achieve robust and high-precision registration. 
This category of methods first utilizes a regressor to initialize the pose, followed by fine-tuning through either differentiable~\cite{gopalakrishnan2022fast} or non-derivative optimization~\cite{li2025automatic}.
Literature~\cite{gopalakrishnan2024intraoperative} indicates that this approach can achieve sub-millimeter registration accuracy within a few seconds in patient-specific scenarios.
\subsection{Similarity Learning for Image Registration}
Image registration is often regarded as a task without a definitive gold standard, making the evaluation of image similarity a critical challenge~\cite{markelj2012review}.
Traditional similarity metrics, such as sum of squared differences (SSD), NCC, and MI, exhibits convexity around the optimum but non-convexity elsewhere, which can trap the estimates into undesirable local minima. 
Similarity learning enables the generation of a globally convex and smooth optimization landscape by allowing neural networks to learn feature representations of images~\cite{chen2024fully,qin2019unsupervised,grzech2022variational}. 
The resulting similarity metric can serve as a differentiable approximation of traditional metrics~\cite{ronchetti2023disa}, a deep distance metric capturing the inherent geometric structure of the image~\cite{grzech2024unsupervised,sideri2023mad}, or a deep neighborhood self-similarity measure that enhances robustness in registration tasks~\cite{mok2024modality}. 
Gu \textit{et al.}\cite{gu2020extended} first introduced similarity learning into the field of 2D/3D registration by proposing a method to learn a convex similarity metric through the estimation of the Riemannian pose gradient. Building on this foundation, Gao \textit{et al.}~\cite{gao2020generalizing,gao2023fully} extended the approach to a fully differentiable framework, while Chen \textit{et al.}~\cite{chen2024embedded} designed a parameter-specific pose initialization strategy to enhance its performance. In contrast to these existing methods, our approach considers the approximation of the distance gradient on the Riemannian manifold in a non-Euclidean space and further extends the learning strategy from SE(3) to SO(4).
This extension ensures that the learned deep metric inherits the bi-invariant properties of SO(4), leading to improved stability and consistency in geometric optimization for 2D/3D registration.
\section{Method}

% \subsection{Preliminaries}
\subsection{Problem formulation}
% \paragraph{Problem formulation.}

2D/3D registration aims to estimate the optimal transformation that aligns a 3D volume (\textit{e.g.}, CT scan) with a 2D projection (\textit{e.g.}, X-ray image) by minimizing a similarity metric. Formally, given a CT volume $V\in \mathbb{R}^3$ in the reference coordinate system and a 2D radiograph $I\in \mathbb{R}^2$, the goal is to find the transformation parameters \( \mathbf{T} \) that align \( \ I \) with \( \ \mathcal{P}(\mathbf{T})\circ V \), where $\mathcal{P}(\cdot)$ is a renderer that generates 2D images from a 3D volume through projection based on medical physics principles, and $\circ$ is warping operation.

The spatial transformation can be represented as a rigid-body transformation in \(\text{SE}(3)\):
\begin{equation}
    \mathbf{T} = 
    \begin{bmatrix}
        \mathbf{R} & \mathbf{t} \\
        0 & 1
    \end{bmatrix} \in \text{SE}(3)
\end{equation}
where \( \mathbf{R} \in \text{SO}(3) \) is the rotation matrix and \( \mathbf{t} \in \mathbb{R}^3 \) is the translation vector.
The registration objective is formulated as an optimization problem:
\begin{equation}
    \mathbf{T}^* = \arg\min_{\mathbf{T} \in \text{SE}(3)} \mathcal{S}(I,\mathcal{P}(\mathbf{T})\circ V)
\end{equation}
where \( \mathcal{S}(\cdot) \) is a similarity function that measures the alignment quality between the transformed 2D projections and their 2D references.
% \paragraph{Preprocessing of real X-ray data.}
% As mentioned in numerous prior studies~\cite{gopalakrishnan2024intraoperative,unberath2018deepdrr,momeni2024voxel}, X-ray imaging captures the intensity attenuation of X-rays as they traverse a medium, whereas DRRs quantify the absorption of X-ray energy by the medium.
% Therefore, in DRRs, higher-density bone tissue appears brighter, while lower-density soft tissue appears relatively darker. This contrast is inverted in X-ray images, where bone structures appear darker due to greater X-ray absorption, while soft tissue appears brighter due to lower attenuation.
% To ensure that the grayscale distribution of the DRR generated using Siddon's method~\cite{siddon1985fast} aligns with that of X-ray images governed by the Beer-Lambert law, we applied the following inversion strategy:
% \begin{equation}
%     \Tilde{I} = 1 - \frac{\log(1+I)}{\log(1+I_0)}
% \end{equation}
% where $I_0$ represents the initial energy of the radiation beam, and all X-ray image pixel values are normalized to their maximum intensity. The transformation $\log(1+I)$ ensures numerical stability by preventing negative logarithmic values while enhancing visual contrast. Additionally, it promotes a Gaussian-like intensity distribution, improving the consistency of the processed images.
\subsection{Network Architecture}
The overall framework of our proposed method is shown in~\cref{fig1}. 
Let $I$ be the set of X-ray images and $V$ be the set of CT volumes. Similar to many recent approaches~\cite{chen2024embedded,gopalakrishnan2024intraoperative,zhang2023patient}, we employ a regressor $\mathcal{R}(\cdot)$ to initialize the pose. Our framework adapts the choice of the regressor based on the application scenario. In patient-agnostic scenarios, the regressor takes both $V$ and $I$ as inputs. However, in patient-specific scenarios, where training is performed exclusively on a subject’s CT and corresponding X-ray images, the regressor operates solely on $I$. This is because patient-specific registration does not require learning the anatomical variability present in the CT volume across different individuals, allowing the model to focus on refining pose estimation based on the subject’s X-ray features.
The regression output $\theta_{ini}$ is parameterized in 
$\mathfrak{se}(3)$, which is isomorphic to $\mathbb{R}^3\times\mathbb{R}^3$ and jointly represents an axis-angle rotation and a translation vector~\cite{murray2017mathematical}. This parameterization has been shown to achieve the highest accuracy in 2D/3D registration~\cite{gopalakrishnan2024intraoperative}.
Next, we map 
$\theta_{ini}$ to the SE(3) to obtain $\textbf{T}_{ini}$, and then use the projector $\mathcal{P}(\cdot)$ to generate the DRR image $I_{ini}$ at the initial pose defined by $\textbf{T}_{ini}$.

Previous study~\cite{gao2023fully} has shown that passing $V$ through a residual 3D CNN before projection can enhance registration accuracy by introducing a learnable parameter module in the 3D domain. We hypothesize that this CNN module captures intensity information to compensate for the domain gap between the projected DRRs and real X-ray images. However, while CNNs can achieve accurate rigid registration, their design does not inherently exploit the symmetry of rigid motion. Convolutional kernels are naturally equivariant to translation, meaning their output shifts with the input, but they lack equivariance to rotation. As a result, CNNs may extract different features for the same object in different poses, making it challenging to learn features that remain consistent under rigid transformations.
To address the limitation above, we employ a Steerable Equivariant CNNs (E-CNNs)~\cite{moyer2021equivariant,billot2024se} $\Psi(\cdot)$ to extract robust SE(3)-equivariant feature maps. A neural network is considered SE(3)-equivariant if, under a rigid body transformation $\textbf{T}$, it satisfies the following condition for an input volume $V$:
\begin{equation}
\Psi(\textbf{T}\circ V)=\textbf{T}\circ \Psi(V)
\end{equation}
% where $\textbf{T}\circ V$ is $V$ warped by $\textbf{T}$.
Under the constraint of the equivariant property, we can ensure a consistent correspondence between $V$ and the output $\Psi(\textbf{T}\circ V)$, preserving geometric relationships across transformations. This alignment enhances feature consistency, thereby stabilizing the learning process.
The output of $V$ and the E-CNNs is combined through element-wise addition to obtain the volume $\hat{V}$, which is enhanced by equivariant learning. The enhanced volume $\hat{V} $ is then projected to generate the DRR image $\hat{I}$. This process can be formally expressed as:
\begin{equation}
    \hat{V}=V + \Psi(V)
\end{equation}
\begin{equation}
    \hat{I}=\mathcal{P}(\mathbf{T})\circ \hat{V}
\end{equation}
The input  $\hat{I}$ and X-ray images $I$ are processed by two networks, $\xi_m$ and $\xi_f$, both of which have identical architectures. Each network consists of a CNN-based branch $\xi^l$ that captures high-frequency texture  details and a Transformer-based branch $\xi^g$ that extracts low-frequency global features. The outputs of these two branches are then concatenated to form the final learned feature representation $\phi$, effectively integrating both local fine-grained details and global contextual information for robust image representation:
\begin{equation}
    \phi_m=\{\xi^l_m(\hat{I}), \xi^g_m(\hat{I})\},  \phi_f=\{\xi^l_f(I), \xi^g_f(I)\}
\end{equation}
The similarity between two images $\varepsilon$ is then obtained by measuring the distance between the two latent vectors $\phi_m$ and $\phi_f$:
\begin{equation}
    \varepsilon = \mathcal{S}(\phi_m,\phi_f)
\end{equation}
where $\mathcal{S}$ is the learned spherical similarity, which will be introduced in section~\ref{section::spherical}.
\subsection{Spherical Similarity Learning}
\label{section::spherical}
To quantify the similarity between two latent vectors 
$\phi_m$ and $\phi_f$, we first embed them onto a unit sphere through a Riemannian mapping from Euclidean space to the spherical manifold. Specifically, let
\begin{equation}
    \Phi_m=\text{EXP}(\phi_m),  \quad \Phi_f=\text{EXP}(\phi_f)
\end{equation}
where $\mathrm{EXP}(\cdot)$ denotes the \textit{spherical exponential map}~\cite{wilson2014spherical}.
This map is defined such that any vector $\phi \in \mathbb{R}^{d}$ is mapped to a point $\Phi$ on the unit sphere $S^d \subset \mathbb{R}^{d+1}$.
We then construct a vector $\bar{\phi}=[\phi,1]\in T_{\textbf{N}}\mathbb{S}^{d}$ in the tangent space of the north pole $\textbf{N}=[0, 0, ...,0,1]\in \mathbb{S}^{d}$
and it can be expressed as:
% \begin{equation}
%     \mathrm{EXP}\bigl(\boldsymbol{\phi}\bigr)
%     =
%     \frac{\bigl(\boldsymbol{\phi},\,\sqrt{1-\|\boldsymbol{\phi}\|^2}\,\bigr)}%
%     {\max\bigl(1,\|\boldsymbol{\phi}\|\bigr)}
% \end{equation}
\begin{equation}
    \mathrm{EXP}\bigl(\boldsymbol{\phi}\bigr)
    =
    \textbf{N}\cos{\left\|\bar{\phi}\right\|}+\bar{\phi}\frac{\sin{\left\|\bar{\phi}\right\|}}{\left\|\bar{\phi}\right\|}
\end{equation}
Once $\Phi_m$ and $\Phi_f$ lie on the sphere, we measure their \textit{spherical distance} by
\begin{equation}
    d\bigl(\Phi_m,\Phi_f\bigr)
    =
    \arccos\Bigl(\langle \Phi_m,\;\Phi_f\rangle\Bigr)
\end{equation}
where $\langle \cdot,\cdot\rangle$ denotes the standard inner product in $\mathbb{R}^{d+1}$. By using this geodesic distance on the sphere, we ensure that the underlying geometry is faithfully preserved.
In practice, the dimension of the embedding vector output by our network is $\in \mathbb{R}^{H\times W\times D}$, and we only perform spherical exponential mapping in the last dimension, \textit{i.e.}, $\Phi[i,j,:]=\text{EXP}(\phi[i,j,:])$.
Finally, we define the \textit{spherical similarity} $S(\boldsymbol{\phi}_m,\boldsymbol{\phi}_f)$ as a (monotonically decreasing) function of the spherical distance. For example, one may choose an exponential decay:
% \begin{equation}
% \label{eq:similarity}
%     \varepsilon 
%     =
%     S\bigl(\boldsymbol{\phi}_m,\boldsymbol{\phi}_f\bigr)
%     =
%     \exp\Bigl(-\alpha\,d\bigl(\Phi_m,\Phi_f\bigr)\Bigr)
% \end{equation}
\begin{equation}
\label{eq:similarity}
    \varepsilon 
    =
    S\bigl(\boldsymbol{\phi}_m,\boldsymbol{\phi}_f\bigr)
    =
    \sum^{H}_{i=1}\sum^{W}_{j=1}(1-\Phi_m[i,j,:]^T\Phi_f[i,j,:])
\end{equation}
% where $\alpha>0$ is a hyperparameter controlling how rapidly the similarity decreases with increasing spherical distance. 
In this way, two feature vectors that are close on the sphere (\textit{i.e. }, have a small geodesic distance) yield a high similarity score, while vectors that lie farther apart yield a lower score.

In this work, we adopt the \emph{double-backward strategy}~\cite{gao2023fully} to approximate the gradient of the network's loss with the gradient of a geodesic distance on the pose manifold. This approach has proven effective in mitigating sensitivity to absolute loss scaling, as it prioritizes preserving the local geometric structure of the manifold rather than relying solely on Euclidean differences.
Specifically, let $\theta \in \mathfrak{se}(3)$ denote the pose of the camera. We denote by $\theta_{gt}$ the corresponding ground-truth pose. Our network produces a similarity score $\mathcal{L}_{\mathrm{net}}=\varepsilon$, whose gradient with respect to $\theta$ is
    $\nabla_{\theta}\,\mathcal{L}_{\mathrm{net}}$.

\noindent
\textbf{Geodesic distance for gradient approximation.} 
We introduce a \emph{geodesic loss} $\mathcal{L}_{\mathrm{geo}}(\theta_A, \theta_B)$ to measure the distance between two poses, \textit{e.g.} $\theta_A$ and $\theta_B$, on Riemannian manifold.
% Recall that any transformation $\textbf{T} \in \mathrm{SE}(3)$ can be represented by a rotation $R \in \mathrm{SO}(3)$ and a translation $\mathbf{t}\in\mathbb{R}^3$. 
One common choice is to measure the pose error on $\mathfrak{se}(3)$~\cite{hou2018computing}, and the geodesic distance on $\mathfrak{se}(3)$ is given by
% \begin{equation}
\begin{align}
\mathcal{L}_{\text{geo}}^{\mathfrak{se}(3)}(\theta_A, \theta_B)
& = \left\| \mathcal{LOG}\Bigl(\textbf{T}^{-1}_A\textbf{T}_B \Bigr)\right\|\\
& = \left\|\mathcal{LOG}\Bigl(\mathcal{EXP}(-\theta_A)\,\mathcal{EXP}(\theta_B)\Bigr)\right\| \notag
\end{align}
% \end{equation}
where $\mathcal{LOG}:\text{SE}(3)\rightarrow \mathfrak{se}(3) $ and $\mathcal{EXP}:\mathfrak{se}(3)\rightarrow \text{SE}(3)$ are the matrix logarithm and exponential mapping between $\mathfrak{se}(3)$ and $\text{SE}(3)$. 
% Equivalently, for the rotation-only case $R_A,R_B\in \mathrm{SO}(3)$, we may use
% \begin{equation}
%     d\bigl(R_A,R_B\bigr) 
%     \;=\; \arccos\!\Bigl(\tfrac{\mathrm{trace}\!\bigl(R_A^\top R_B\bigr) - 1}{2}\Bigr)
%     \quad\text{or}\quad
%     \bigl\|\log\bigl(R_A^\top R_B\bigr)\bigr\|.
% \end{equation}
This geodesic definition captures the angular distance on the manifold, rather than an absolute Euclidean difference~\cite{mahendran20173d,salehi2018real}.
% Given a 3D pose represented in the special Euclidean group \(\mathrm{SE}(3)\), we denote the pose by
% \begin{equation}
%     \label{eq:se3_T}
%     T = \begin{bmatrix}
%            \mathbf{R} & \mathbf{t} \\
%            \mathbf{0}^\top & 1
%         \end{bmatrix}, \quad \mathbf{R} \in \mathrm{SO}(3), \quad \mathbf{t} \in \mathbb{R}^3.
% \end{equation}
However, since the geodesic distance on  SE(3) is only left-invariant, the learned similarity function may exhibit ambiguity and inconsistency in its optimization landscape. To address this issue, we explore approximating the gradient within the bi-invariant structure of SO(4), which provides a more stable and geometrically consistent formulation.
Following established approaches~\cite{meng2025parametric,thomas2014approaching}, the mapping from $\theta \in \mathfrak{se}(3) \rightarrow \textbf{T} \in \mathrm{SE}(3)$ to \(\textbf{H} \in \mathrm{SO}(4)\) is defined as:
\begin{equation}
    \label{eq:so4_mapping}
    \textbf{H} =\mathcal{M}(\theta) =\begin{bmatrix}
           \mathbf{R} & \dfrac{\mathbf{t}}{2f} \\
           -\mathbf{t}^\top \mathbf{R} & 1
        \end{bmatrix}
\end{equation}
where \(f\) is the source-to-detector radius of the C-arm system. 
% The block structure in Eq.~\eqref{eq:so4_mapping} guarantees that \(\textbf{H}\) is an element of \(\mathrm{SO}(4)\); that is, \(\textbf{H}\) is orthogonal (\(\textbf{H}^\top \textbf{H} = I_4\)) and \(\det(\textbf{H})=+1\).
% \subsection{Geodesic Loss on \texorpdfstring{\(\mathrm{SO}(4)\)}{SO(4)}}
Once the pose is embedded in \(\mathrm{SO}(4)\) via \(\textbf{H}\), we can define a geodesic loss that measures the discrepancy between a predicted pose and the ground-truth pose in a unified framework.
Given two poses \(\textbf{H}_A, \textbf{H}_B \in \mathrm{SO}(4)\), obtained by mapping the corresponding  \(\mathrm{SE}(3)\) representations through~\cref{eq:so4_mapping}, the geodesic distance can be expressed using the Frobenius norm as:
% For two poses \(\textbf{H}_A, \textbf{H}_B \in \mathrm{SO}(4)\) (obtained by mapping the corresponding \(\mathrm{SE}(3)\) poses via Eq.~\eqref{eq:so4_mapping}), one may define a geodesic distance using the Frobenius norm:
\begin{equation}
    \label{eq:geo_loss_so4_matrix}
    \mathcal{L}_{\mathrm{geo}}^{\mathrm{SO(4)}}(\theta_A, \theta_B) =
    \|\textbf{H}_A - \textbf{H}_B\|_F=\|\mathcal{M}(\theta_A), \mathcal{M}(\theta_B)\|_F
\end{equation}
To avoid the impact of absolute loss scale, we follow the idea of previous works~\cite{gao2023fully,chen2024fully} and approximate the gradient of $\mathcal{L}_{\mathrm{net}}$ by aligning it with the gradient of the geodesic distance. Concretely, we want:
$    \nabla_{\theta}\,\mathcal{L}_{\mathrm{net}}(\theta)
    \;\approx\;
    \nabla_{\theta}\,\mathcal{L}_{\mathrm{geo}}\!\bigl(\theta,\theta_{gt}\bigr)$.
We thus define an \emph{approximate loss} $\mathcal{L}$ that measures the mismatch between these two gradients:
\begin{equation}
    \label{eq:double-backward}
    \mathcal{L}
    \;=\;
    \mathcal{L}_{\mathrm{geo}}\Bigl(
        \tfrac{\partial \mathcal{L}_{\mathrm{net}}(\theta)}{\partial \theta},
        \tfrac{\partial \mathcal{L}_{\mathrm{geo}}(\theta,\theta_{gt})}{\partial \theta}
    \Bigr)
\end{equation}
% Here, $\mathcal{L}_{\mathrm{geo}}(\cdot,\cdot)$ can again be interpreted as a distance measure on the chosen manifold of gradients (e.g., a spherical distance if we normalize gradients onto a unit sphere, or directly a geodesic distance if we embed them into an appropriate Lie group representation).

% By minimizing $\mathcal{L}_{\mathrm{appro}}(\theta,\theta_t)$, we encourage
% \begin{equation}
%     \frac{\partial \mathcal{L}_{\mathrm{net}}(\theta)}{\partial \theta}
%     \;\longrightarrow\;
%     \frac{\partial \mathcal{L}_{\mathrm{geo}}(\theta,\theta_t)}{\partial \theta},
% \end{equation}
% thus aligning the network's gradient with that of the geodesic distance. This \emph{double-backward strategy} has proven effective in mitigating the sensitivity to absolute loss scaling, as it focuses on preserving the local geometric structure in $\mathfrak{se}(3)$ (or in a spherical embedding) rather than purely Euclidean differences.

% \vspace{0.5em}
% \noindent
% \textbf{Implementation notes.} 
% In practice, one can choose different parameterizations for $\theta$. For instance, if the rotation is represented by quaternions, an alternative geodesic distance can be defined on $\mathrm{SO}(4)$ or directly on the 4D hypersphere $S^3$. The key idea remains to \emph{compute a manifold-aware distance} and \emph{align its gradient} with that of the network's primary loss. This ensures that the pose regression respects the underlying geometry of rotations and translations, improving stability and accuracy in 3D vision tasks.

\subsection{Training and Inference}
Given the challenges of 2D/3D registration data annotation and collection, we adopt a self-supervised training strategy, similar to other works~\cite{gao2020generalizing,gopalakrishnan2024intraoperative}. In each iteration, a ground truth pose $\theta_{gt}$ is randomly generated near the patient’s isocenter, and a DRR is synthesized as image $I$ using a projector $\mathcal{P}(\cdot)$. 
The training process is divided into two stages. For the regressor  $\mathcal{R}(\cdot)$ for pose initialization, we directly compute the geodesic distance in $\mathfrak{se}(3)$ between its estimation $\theta_{ini}$ and the ground truth $\theta_{gt}$, \textit{i.e.} $\mathcal{L}_{\text{geo}}^{\mathfrak{se}(3)}(\theta_{ini}, \theta_{gt})\ $. 
The subsequent training of the spherical similarity learning network is guided by~\cref{eq:double-backward}. To further mitigate the domain gap between DRRs and real X-ray images, we apply a domain randomization strategy which is similar to that in~\cite{jaganathan2023self}.

During the inference phase, the estimated initial pose is first obtained through the regressor, followed by differentiable Levenberg-Marquardt (LM) optimization based on spherical similarity learning.
In the $i$-th iteration of the optimization process, the pose parameter is initialized with the estimated pose from the previous iteration, \textit{i.e.},  $\theta_0=\theta_{ini}$.
The pose refinement is then formulated as an optimization problem:
$\hat{\theta}=\arg\min_{\theta}\mathcal{L}_{net}(\theta)$.
% \begin{equation}
%     \hat{\theta}=\arg\min_{\theta}\mathcal{L}_{net}(\theta)
% \end{equation}
At each LM iteration, starting from the previous estimate $\theta_{i-1}$, the left-multiplied pose increment $\Delta\theta_t$ is computed as
\begin{equation}
\Delta \theta_i = (J^T W J + \lambda I)^{-1} J^T W \mathbf{r}(\theta_{i-1})
\end{equation}
where $J = - \frac{\partial \mathbf{r}}{\partial \theta}$ is the Jacobian matrix capturing the derivatives of the stacked residuals with respect to the left-multiplied pose increment, $W$ is the weight matrix, and $\mathbf{r}(\theta_{i-1})$ represents the residual vector, derived from~\cref{eq:similarity}.
Finally, the pose update is performed as $ \theta_i \leftarrow \Delta \theta_i \circ \theta_{i-1}$, iteratively refining the pose to converge toward the optimal solution.
% \section{Final copy}

% You must include your signed IEEE copyright release form when you submit your finished paper.
% We MUST have this form before your paper can be published in the proceedings.

% Please direct any questions to the production editor in charge of these proceedings at the IEEE Computer Society Press:
% \url{https://www.computer.org/about/contact}.
\begin{figure*}[h!]
\centering
\includegraphics[width=\linewidth]{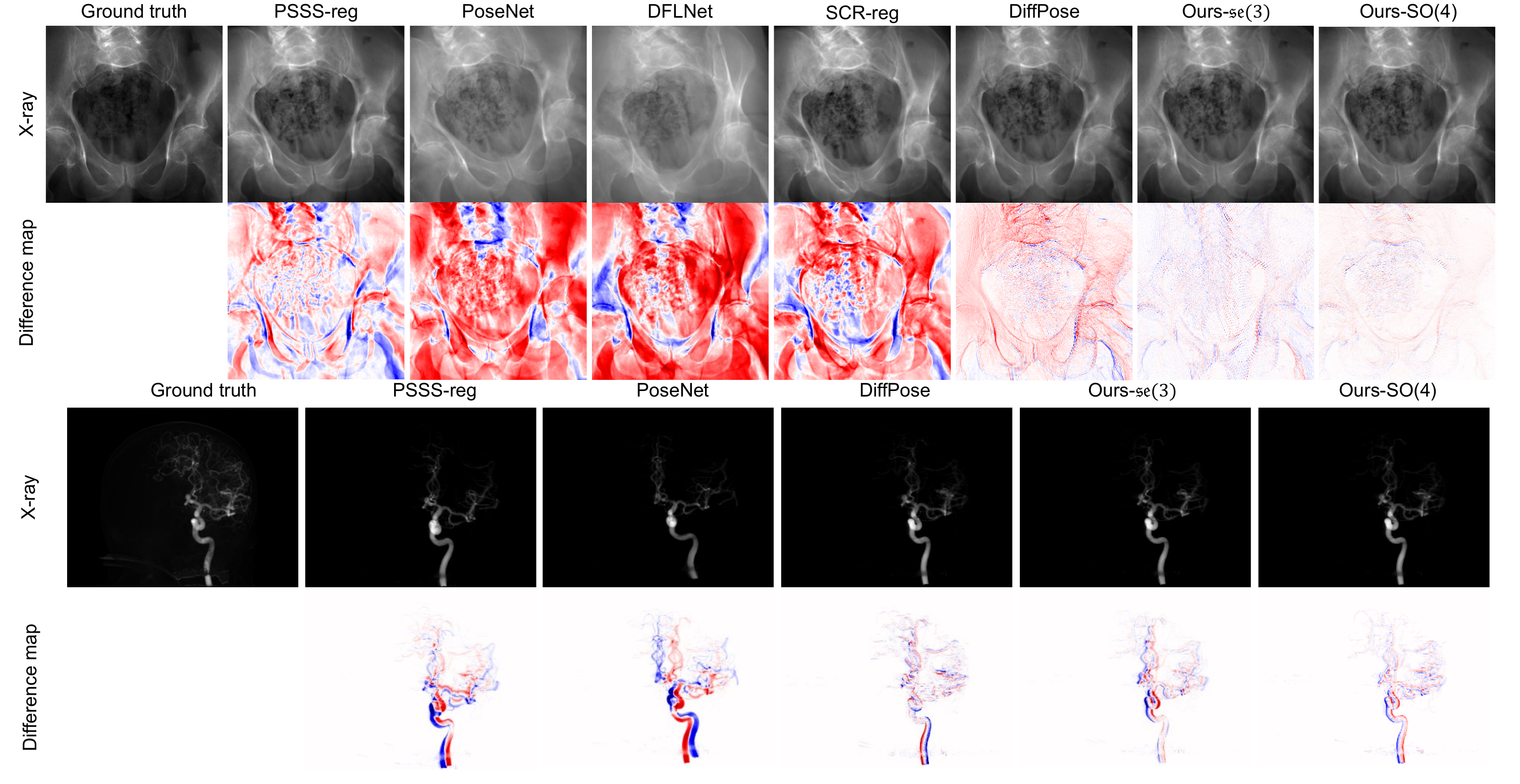}
\caption{\textbf{Qualitative results.}
    We visualize the registration results of the proposed method and the baselines on the DeepFluoro 
 and Ljubljana dataset under the patient-specific scenario. In each example, the \textbf{top} row shows the ground truth X-ray image alongside the corresponding DRR generated from the pose estimated by each method. The \textbf{bottom} row displays the difference map, demonstrating the alignment between the DRR at the ground truth pose (\textcolor{red}{red}) and the DRR at the estimated pose (\textcolor{blue}{blue}).
}
\label{fig::qualitative}
\end{figure*}
\section{Experiments}
\subsection{Experimental Settings}
We perform experiments on three publicly available datasets to evaluate the effectiveness of the proposed method:  DeepFluoro~\cite{grupp2020automatic}, Ljubljana~\cite{pernus20133d}, and CTSpine1K~\cite{deng2021ctspine1k}.
The details of the real X-ray data preprocessing and these three datasets mentioned above can be found in \textbf{supplementary material} ~\cref{sec::preprocess} and \cref{sec::dataset} respectively.
We evaluated the proposed method in two application scenarios: patient-specific and patient-agnostic ( \textit{i.e.} generic 2D/3D registration).
In the patient-specific experiment, we performed self-supervised training on each subject individually, followed by registration with its corresponding X-ray image during the test phase. The final performance was reported as the average across all subjects.
For the patient-agnostic scenario, we partitioned the dataset into training, validation, and test sets based on subjects, using a 7:2:1 split (and 4:1:1 for DeepFluoro). The evaluation was conducted on the test set, and the results were reported accordingly. Since achieving high registration accuracy solely with learning-based methods is challenging in patient-agnostic scenarios, we incorporated an optimization-based approach~\cite{chen2024optimization,chen2025introducing} to refine the final result, enhancing the overall alignment precision.

The evaluation metrics used in our experiments are as follows:
\begin{itemize}
    \item \textbf{Mean target registration error (mTRE)}: This metric measures the average distance between corresponding landmarks in the warped image and the target image, reflecting the accuracy of alignment. We report the alignment performance at the 50th, 75th, and 95th percentiles to provide a comprehensive evaluation of registration quality. For the CTSpine1K dataset, landmarks are obtained by extracting contour points along the surface of the spine, ensuring anatomically relevant feature correspondence.

    \item \textbf{Sub-millimeter success rate (SMSR)}: We define a successful registration as achieving a mTRE $<$ 1 mm, ensuring high-precision alignment for clinical applicability.
\end{itemize}
In patient-agnostic scenarios, we compared the proposed method with open-source learning-based baselines, including ProST-m~\cite{gao2020generalizing}, ProST-t~\cite{gao2023fully}, SOPI~\cite{chen2024embedded} and CDreg~\cite{chen2024fully}, as well as a traditional optimization-based method that utilizes BOBYQA~\cite{powell2009bobyqa} as the optimizer and multi-scale normalized cross-correlation~\cite{gopalakrishnan2024intraoperative} as the similarity measure. To ensure a fair comparison, all three learning-based baselines also underwent the same refinement process as the proposed method and used the same pose initialization network.
As the smaller sample sizes in the other two datasets limited their suitability for this analysis, we conducted experiments exclusively on the CTSpine1K dataset, which has the largest number of subjects.
In the patient-specific scenario, our baseline methods include PSSS-reg~\cite{zhang2023patient}, DiffPose~\cite{gopalakrishnan2024intraoperative}, PoseNet~\cite{bui2017x}, DFLNet~\cite{grupp2020automatic} and SCR-reg~\cite{shrestha2023x}. We follow the original parameter settings of other methods except for an additional grid search for the hyperparameters of SCR-reg.
To be note that, since the Ljubljana dataset lacks landmark annotations, landmark-based methods (DFLNet and SCR-reg) cannot be applied to this dataset.
The implementation details of the proposed method is provided in \textbf{supplementary material}~\cref{sec::imple}
% \subsection{Implementation Details}
% For patient-specific 2D/3D registration, we selected ResNet-18 as the backbone for the pose regressor $\mathcal{R}(\cdot)$, using the Adam optimizer with a learning rate of 1e-3 and a cosine scheduler. The spherical similarity network, used for subsequent fine-tuning, was trained with the AdamW optimizer, also with a learning rate of 1e-3, but utilizing a cyclic scheduler. The pose regressor and similarity network were trained in a self-supervised manner using 500k and 200k synthetic images generated in real-time, respectively.
% For the patient-agnostic scenario, the parameter settings remained largely the same, except that the pose regressor was replaced with RTPIv3~\cite{chen2024embedded}, which incorporates spatial information from CT. Additionally, the number of self-supervised training iterations was increased to 750k for the pose regressor and 300k for the similarity network, to better generalize across different subjects.
% The termination criterion for the differentiable LM optimization is set such that the standard deviation of $\Delta\theta_i$ over the last ten iterations is less than 1e-2. 
% This threshold was determined through grid search across all datasets to ensure an optimal balance between convergence stability and computational efficiency.
% Our experiments were conducted on a PC equipped with an NVIDIA RTX 3090 GPU and an Intel Core i7 CPU. Due to memory limitations, the batch sizes for training the pose regressor and the spherical similarity network were set to 4 and 1, respectively.
\subsection{Results}

\begin{figure}[h!]
    \centering
    \begin{subfigure}[b]{0.48\textwidth}
        \centering
        \includegraphics[width=\textwidth]{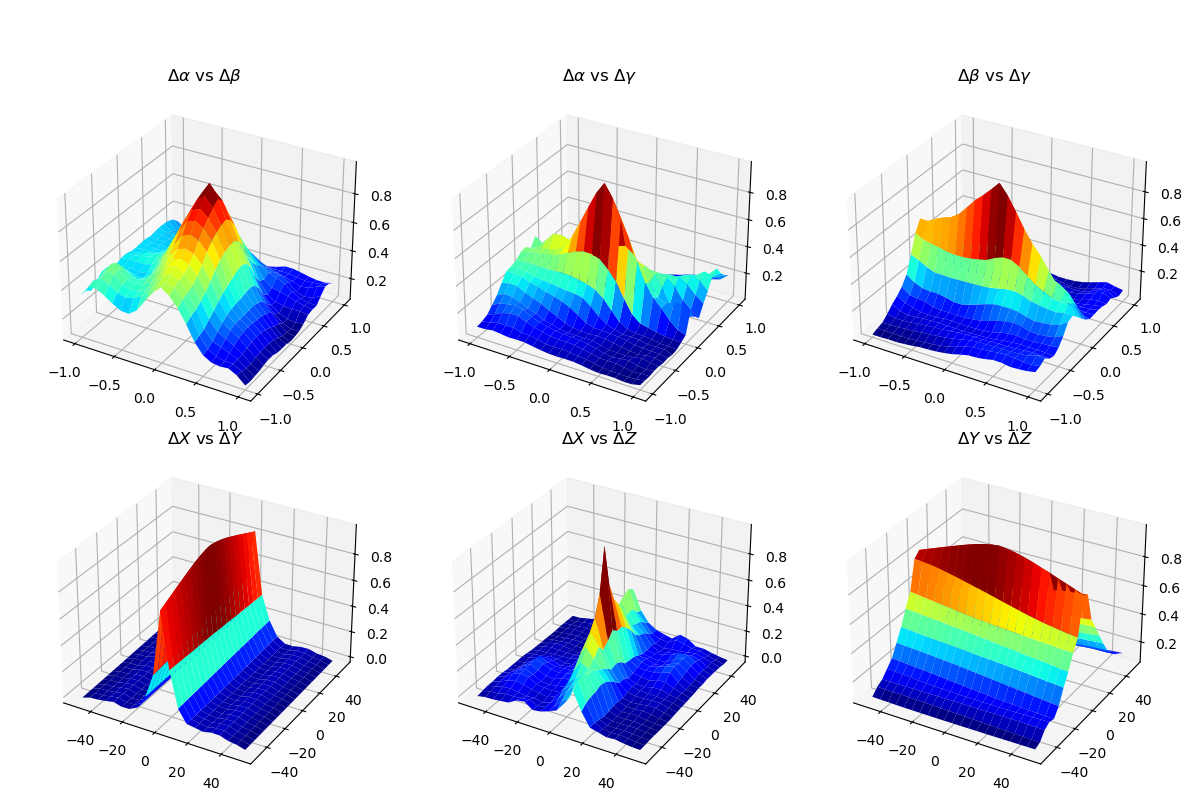}
        % \caption{Subfigure 1}
    \end{subfigure}
    \begin{subfigure}[b]{0.48\textwidth}
        \centering
        \includegraphics[width=\textwidth]{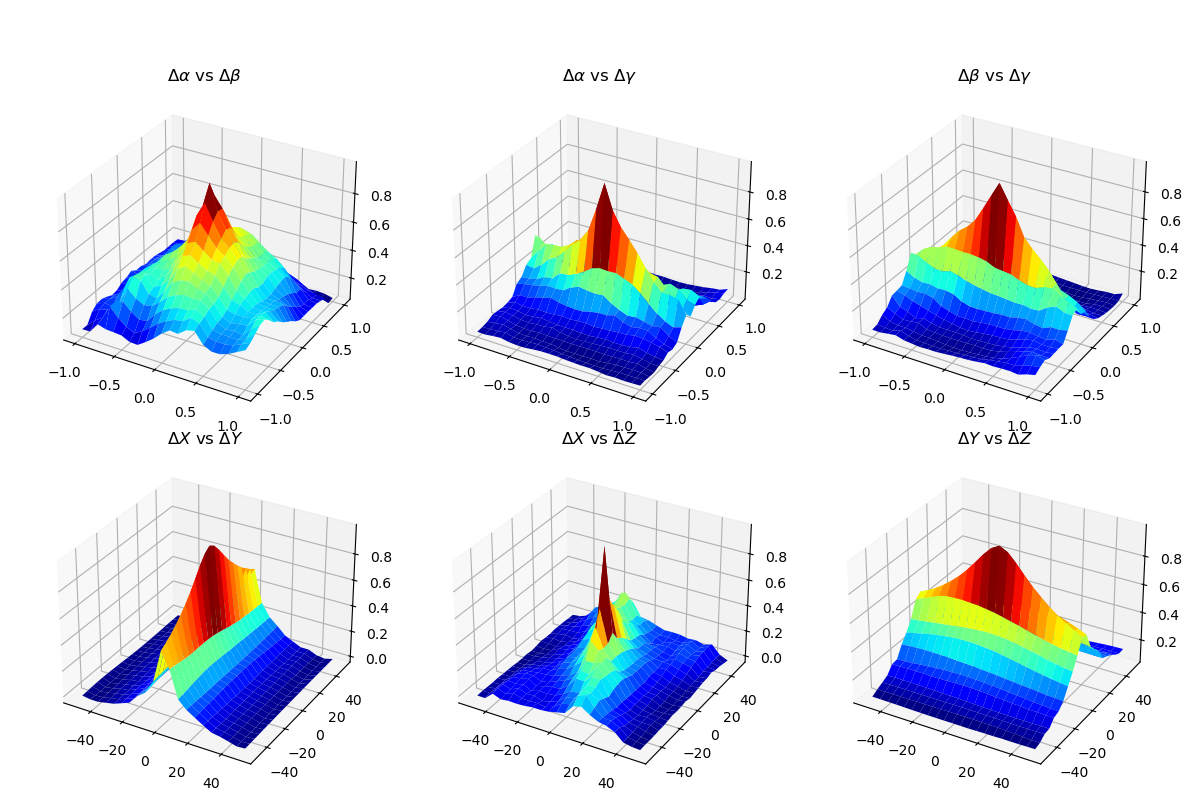}
        % \caption{Subfigure 2}
    \end{subfigure}
\caption{Visual comparison of the proposed spherical deep similarity landscape in  $\mathfrak{se}(3)$ (top) and SO(4) (bottom).
For clearer visualization, the deep similarity values are first normalized to the range [0,1], and then transformed by computing 1-$\epsilon$, effectively inverting the scale to enhance contrast in the display.}
\label{fig::similarity}
\end{figure}

\begin{table}[h]
    \centering
    \caption{Experiment results on patient-specific 2D/3D registration on 366 test cases from 20 pelvic CTs, 20 test cases from 10 clinical intraoperative CBCTs, and 502,500 test cases from 1005 spine CTs. Sub-millimeter success rate (SMSR) accounts for test cases with mTRE$<$1 mm. Median, 75th percentile and 95th percentile mTREs are reported.
    The best results are \textbf{bolded}.}
    \resizebox{\linewidth}{!}{
    \begin{tabular}{c l c c c c c}
        \toprule
        \multirow{2}{*}{} & \multirow{2}{*}{\textbf{Method}} & \multirow{2}{*}{\textbf{SMSR}} & \textbf{Median} & \multicolumn{2}{c}{\textbf{Percentile (mm)}} & \textbf{Run} \\
        \cmidrule(lr){5-6}
        & & & (mm) & 75\% & 95\% & \textbf{Time} \\
        \midrule
        \multirow{6}{*}{\rotatebox{90}{\textbf{DeepFluoro}}} 
        & PSSS-reg~\cite{zhang2023patient}         & 56.0\%  & 0.93  & 2.51  & 5.57  & 12.7 s   \\
        & PoseNet~\cite{bui2017x}       & 4.3\%   & 16.6  & 22.0  & 29.2  & \textbf{0.1 s}  \\
        & DFLNet~\cite{grupp2020automatic}      & 36.6\%   & 3.20  & 7.29  & 13.1  & 1.0 s \\
        & SCR-reg~\cite{shrestha2023x}      & 33.3\%  & 4.70  & 9.59  & 12.8  & 1.1 s \\
        & DiffPose~\cite{gopalakrishnan2024intraoperative}     & 83.1\%   & 0.60  & 0.89  & 1.47  & 5.3 s\\
          \cmidrule{2-7}
        & Ours-$\mathfrak{se}(3)$   & 82.8\%   & 0.60  & 0.89  & 1.77  & 5.6 s \\
        & Ours-SO(4)   & \textbf{86.1\%}   & \textbf{0.51}  & \textbf{0.85} & \textbf{1.42}  & 6.2 s \\
        \midrule
        \multirow{6}{*}{\rotatebox{90}{\textbf{Ljubljana}}} 
         & PSSS-reg~\cite{zhang2023patient}         & 40.0\%  & 2.48  & 5.87  & 11.3  & 15.3  s   \\
        & PoseNet~\cite{bui2017x}       & 0\%  & 23.3  & 26.2  & 29.2  & \textbf{$<$0.1 s} \\
        % & DFLNet~\cite{grupp2020automatic}      & 34.3\%  & 1.81  & 21.7  & 41.8  & 17.2 s \\
        % & SCR-reg~\cite{shrestha2023x}      & 17.2\%  & 3.10  & 8.13  & 23.3  & 0.5 s \\
        & DiffPose~\cite{gopalakrishnan2024intraoperative}     & 80.0\%   & 0.63  & 0.94  & 1.78  & 6.0 s \\
          \cmidrule{2-7}
        & Ours-$\mathfrak{se}(3)$   & \textbf{85.0\%}   & 0.57  & \textbf{0.85}  & 1.77  & 6.6 s \\
        & Ours-SO(4)   & \textbf{85.0\%}   & \textbf{0.55}  & \textbf{0.85}  & \textbf{1.35}  & 6.5 s \\
        \midrule
        \multirow{6}{*}{\rotatebox{90}{\textbf{CTSpine1k}}} 
         & PSSS-reg~\cite{zhang2023patient}         & 31.4\%  & 4.57  & 9.75  & 15.8  & 16.2 s   \\
        & PoseNet~\cite{bui2017x}       & 9.8\%  & 11.7  & 17.2  & 24.5  &  \textbf{$<$ 0.1 s} \\
        & DFLNet~\cite{grupp2020automatic}      & 28.4\%  & 4.80  & 10.9  & 18.1  & 1.3 s \\
        & SCR-reg~\cite{shrestha2023x}      & 22.2\%  & 7.08  & 12.6  & 19.7  & 1.3 s \\
        & DiffPose~\cite{gopalakrishnan2024intraoperative}     & 66.4\%   & 0.77  & 1.51  & 3.39  & 7.3 s \\
        \cmidrule{2-7}
        & Ours-$\mathfrak{se}(3)$   & 76.5\%   & 0.65  & 0.97  & 2.11  & 6.6 s \\
        & Ours-SO(4)   & \textbf{80.6\%}   & \textbf{0.59}  & \textbf{0.93}  & \textbf{1.83}  & 6.6 s \\
        \bottomrule
    \end{tabular}
    \label{tab::patient-specific}
    }
\end{table}
As shown in~\cref{tab::patient-specific}, PoseNet's SMSR is very low across all three datasets, indicating that direct regression methods struggle to accurately estimate intraoperative pose. This highlights the limitations of purely regression-based approaches in achieving precise alignment without further refinement.
The registration success rate of landmark-based methods, DFLNet and SCR-reg, is only around 30\%, with most failures occurring due to the limited number of visible landmarks in the 2D images.
In addition, this type of method can only be used when the volume data has pre-annotated landmarks, which reflects its heavy dependence on landmarks.
Among the three iterative methods, DiffPose outperforms PSSS-reg in both accuracy and running time. We attribute this improvement to two key factors: the similarity function mNCC used by DiffPose is better than naïve NCC, and the sparse rendering acceleration it uses also greatly reduces the time required for each iteration.
Our method outperforms all baseline methods across the three datasets, achieving a 15\% improvement over the second-best method on the CTSpine1K dataset. Additionally, the approach based on SO(4) geodesic distance demonstrates higher accuracy compared to the approximation using $\mathfrak{se}(3)$ geodesic distance,  which shows that bi-invariant parameterizations can better enable the network to capture the differences between poses and thus  learn to better represent the inter-image dissimilarity.
% \begin{figure}[h!]
% \centering
% \includegraphics[width=\linewidth]{similarity_1.png}
% \label{fig::similarity}
% \end{figure}
% \begin{figure}[h!]
% \centering
% \includegraphics[width=\linewidth]{similarity_2.png}
% \caption{Visual comparison of the proposed spherical deep similarity landscape in  $\mathfrak{se}(3)$ (top) and SO(4) (bottom).For clearer visualization, the deep similarity values are first normalized to the range [0,1], and then transformed by computing 1-$\epsilon$, effectively inverting the scale to enhance contrast in the display.}
% \label{fig::similarity}
% \end{figure}
\begin{table}[h]
    \centering
    \caption{Experiment results on patient-agnostic 2D/3D registration on CTSpine1k dataset. The best results are \textbf{bolded}.}
     \resizebox{\linewidth}{!}{
    \begin{tabular}{lcccccc}
        \toprule
        \textbf{Method} & \textbf{SMSR} & \textbf{Median} & \multicolumn{2}{c}{\textbf{Percentile (mm)}} & \textbf{Run} \\
        \cmidrule(lr){4-5}
        & & (mm) & 75\% & 95\% & \textbf{Time} \\
        \midrule

        BOBYQA & 18.5\% & 5.01 & 8.02 & 32.4 & 22.3 s \\
        ProST-m~\cite{gao2020generalizing} & 37.6\% & 3.03 & 7.36 & 13.6 & 18.7 s \\
         ProST-t~\cite{gao2023fully} & 46.3\% & 2.03 & 9.56 & 20.4 & 13.2 s \\
        SOPI~\cite{chen2024embedded} & 43.8\% & 1.99 & 6.28 & 12.5 & 14.2 s \\
        CDreg~\cite{chen2024fully} & 50.1\%   & 0.99  & 7.72  & 17.4  & \textbf{10.1 s} \\
        \midrule
         Ours-$\mathfrak{se}(3)$   & 
        53.1\% & 0.94 & 5.01 & \textbf{11.4} & 11.7 s \\
        Ours-SO(4)   & \textbf{55.5\%}   & \textbf{0.90}  & \textbf{4.85}  & 11.9  & 12.5 s \\
        \bottomrule
    \end{tabular}
    \label{tab::non-patient-specific}
    }
\end{table}
The patient-agnostic registration results are presented in~\cref{tab::non-patient-specific}.
Compared to the patient-specific scenario, the registration success rate in this scenario is generally lower because there is no fixed patient anatomical structure as a reference to provide additional spatial information.
However, the key advantage of this approach is that it eliminates the need to train a new model from scratch for each patient, offering a more generalizable and computationally efficient solution.
Compared with the baselines, our method performs better under both pose parameterization schemes, indicating that the architecture of our approach is more effective in optimizing the pose representations derived from image features.
The intensity-based method, BOBYQA, exhibits large errors when the initial pose is far from the ground truth, with the 95th percentile of mTRE reaching 32.4 mm. This is primarily due to the limitations of the handcrafted similarity function it relies on, which results in a narrow capture range and reduced robustness to large initial misalignments.
CDreg has a shorter registration time than our method, but has larger errors than our method at the 75th and 95th percentiles of mTRE, which indicates that although it can converge faster, it also more likely fails at local minima and cause registration failure.
\Cref{fig::qualitative} shows the qualitative results of patient-specific registration between the proposed method and the baselines.
In~\cref{fig::similarity}, we visualize the learned deep spherical similarity landscape parameterized in SO(4) and compare it with the landscape in $\mathfrak{se}$(3). The results show that the similarity derived from approximating the geodesic distance in SO(4) space is not only more globally convex, but also more effectively captures translation variations along the Y-axis, demonstrating improved sensitivity and geometric consistency.
\subsection{Ablation Studies}
\begin{figure}[h!]
  \centering
  % 左子图
  \begin{subfigure}[b]{0.23\textwidth}
    \centering
    \includegraphics[width=\textwidth]{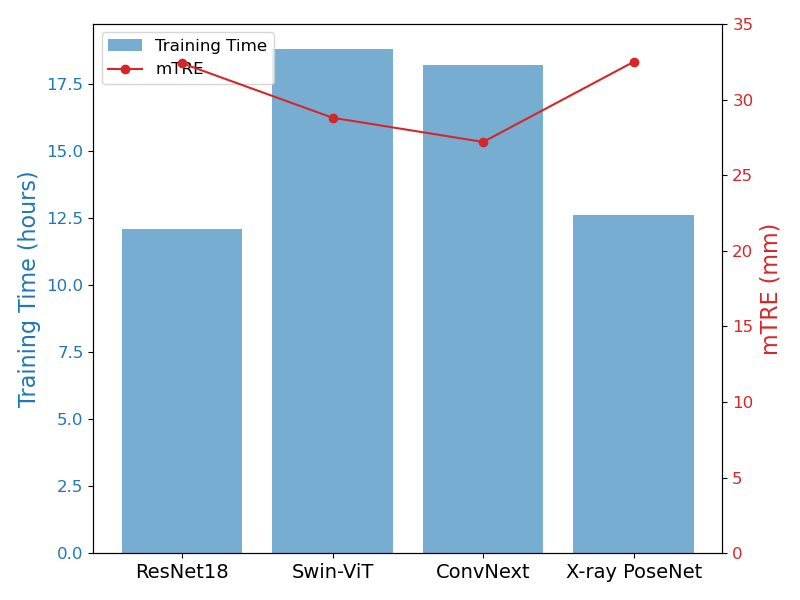}
    % 如果需要子图标题，可用 \caption{...}
  \end{subfigure}
  % 右子图
  \begin{subfigure}[b]{0.23\textwidth}
    \centering
    \includegraphics[width=\textwidth]{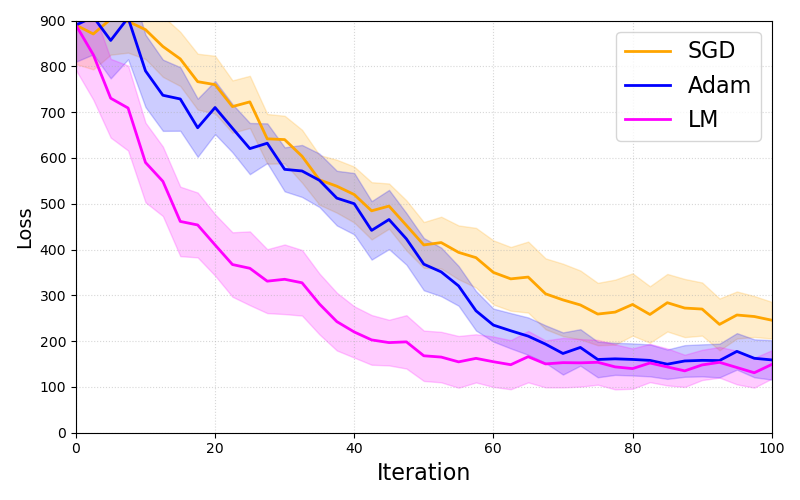}
    % \caption{...}
  \end{subfigure}
  \caption{\textbf{Left:} comparison of training time and mTRE for position regressors with different backbone architectures.The training time reports the time when the standard deviation of the model's loss function is less than 10e-4 in the last ten epochs.
  \textbf{Right:} comparison of the convergence speed of the proposed framework using different gradient-based optimization methods in the inference phase.}
  \label{fig:common}
\end{figure}

\begin{table}[h!]
    \centering
    \caption{Ablation studies of the proposed method on CTSpine1k dataset.}
    \begin{tabular}{lcc}
        \toprule
        & \textbf{SMSR} $\uparrow$ & \textbf{mTRE (mm)} $\downarrow$ \\
        \midrule
        Ours-$\mathfrak{se}(3)$ & 53.1\% & 3.2 $\pm$ 3.9 \\
        Ours-SO(4) & \textbf{55.5\%} & \textbf{2.1 $\pm$ 4.1} \\
        \midrule
       Hyperbolic similarity & 51.3\% & 3.5 $\pm$ 4.3 \\
        Euclidean similarity & 52.8\% & 3.2 $\pm$ 3.9 \\
        w/o E-CNN & 48.2\% & 5.3 $\pm$ 6.7 \\
        w/  3D CNN & 49.9\% & 3.6 $\pm$ 4.4 \\
        \bottomrule
    \end{tabular}
    \label{tab::ablation}
\end{table}
We performed an ablation study on the proposed method on the CTSpine1k dataset. By replacing/removing E-CNN and performing deep similarity learning in different spaces, we verified the effectiveness of each component.
The results are shown in~\cref{tab::ablation}.
Experimental results show that, compared to both Euclidean and other non-Euclidean spaces, similarity learning in spherical space achieves the best approximation performance.
Next, we further investigate the model selection for the pose regressor. In previous experiments, ResNet-18 was used as the backbone. In~\cref{fig:common}(left), we present the registration error and corresponding training time for patient-specific pose regression using several classic network architectures. The results show that models with a larger number of parameters generally achieve lower registration errors but at the cost of significantly increased training time.
However, training times exceeding 12 hours are impractical for the clinical application of this task, so we need to make a trade-off between registration error and training time.  
Given that the subsequent pose refinement stage can reliably guide the solution toward optimal alignment, we conclude that a lightweight backbone is sufficient for pose initialization.
We also compared different optimization methods used in the inference phase of the pose refinement module. As shown in~\cref{fig:common}(right), the differentiable Levenberg-Marquardt optimizer demonstrates faster convergence compared to both SGD and Adam, and achieves a significantly lower loss than SGD. This highlights its effectiveness for fine-tuning pose estimates during registration.
\section{Conclusion}
In this paper, we propose an intraoperative single-view 2D/3D registration method applicable to both patient-specific and patient-agnostic scenarios. Our approach leverages spherical similarity learning to develop a deep similarity metric that effectively captures pose differences based on the anatomical information within the images. During inference, we refine the pose using a differentiable LM optimization algorithm, ensuring precise alignment. We evaluate our method on three datasets covering different human anatomical regions, demonstrating its effectiveness and robustness in various registration tasks.
The main drawback of the existing framework is that the pose regressor and the learned similarity are trained separately, making it time-consuming and labor-intensive to train or retrain a framework for clinical use, thus limiting its practical application.
Future work will focus on improving the training strategy to reduce the time required to train the model, which is crucial for the application of the proposed method to intraoperative navigation systems in emergency situations.
Specifically, our next goal is to integrate patient-specific and patient-agnostic registration approaches by fine-tuning a pre-trained patient-agnostic model to adapt to specific patients. This strategy has the potential to significantly reduce computational overhead.
% In addition, a universal registration model that is agnostic to anatomical structures will also be a further exploration.
\newpage
{
    \small
    \bibliographystyle{ieeenat_fullname}
    \bibliography{main}

@String(CVPR= {IEEE Conf. Comput. Vis. Pattern Recog.})

@String(ICASSP=	{ICASSP})

@String(AAAI = {AAAI})

@String(CVPR  = {CVPR})

@article{unberath2021impact,
  title={The impact of machine learning on 2d/3d registration for image-guided interventions: A systematic review and perspective},
  author={Unberath, Mathias and Gao, Cong and Hu, Yicheng and Judish, Max and Taylor, Russell H and Armand, Mehran and Grupp, Robert},
  journal={Frontiers in Robotics and AI},
  volume={8},
  pages={716007},
  year={2021},
  publisher={Frontiers Media SA}
}

@article{markelj2012review,
  title={A review of 3D/2D registration methods for image-guided interventions},
  author={Markelj, Primoz and Toma{\v{z}}evi{\v{c}}, Dejan and Likar, Bostjan and Pernu{\v{s}}, Franjo},
  journal={Medical image analysis},
  volume={16},
  number={3},
  pages={642--661},
  year={2012},
  publisher={Elsevier}
}

@article{hansen2001completely,
  title={Completely derandomized self-adaptation in evolution strategies},
  author={Hansen, Nikolaus and Ostermeier, Andreas},
  journal={Evolutionary computation},
  volume={9},
  number={2},
  pages={159--195},
  year={2001},
  publisher={MIT Press}
}

@inproceedings{gao2020generalizing,
  title={Generalizing spatial transformers to projective geometry with applications to 2D/3D registration},
  author={Gao, Cong and Liu, Xingtong and Gu, Wenhao and Killeen, Benjamin and Armand, Mehran and Taylor, Russell and Unberath, Mathias},
  booktitle={International Conference on Medical Image Computing and Computer Assisted Intervention},
  pages={329--339},
  year={2020},
  organization={Springer}
}

@article{gao2023fully,
  title={A Fully Differentiable Framework for 2D/3D Registration and the Projective Spatial Transformers},
  author={Gao, Cong and Feng, Anqi and Liu, Xingtong and Taylor, Russell H and Armand, Mehran and Unberath, Mathias},
  journal={IEEE transactions on medical imaging},
  year={2023},
  publisher={IEEE}
}

@inproceedings{miao2016real,
  title={Real-time 2D/3D registration via CNN regression},
  author={Miao, Shun and Wang, Z Jane and Zheng, Yefeng and Liao, Rui},
  booktitle={2016 IEEE 13th International Symposium on Biomedical Imaging (ISBI)},
  pages={1430--1434},
  year={2016},
  organization={IEEE}
}

@inproceedings{guo2021end,
  title={End-to-end ultrasound frame to volume registration},
  author={Guo, Hengtao and Xu, Xuanang and Xu, Sheng and Wood, Bradford J and Yan, Pingkun},
  booktitle={International Conference on Medical Image Computing and Computer Assisted Intervention},
  pages={56--65},
  year={2021},
  organization={Springer}
}

@article{li2012robust,
  title={A robust O (n) solution to the perspective-n-point problem},
  author={Li, Shiqi and Xu, Chi and Xie, Ming},
  journal={IEEE transactions on pattern analysis and machine intelligence},
  volume={34},
  number={7},
  pages={1444--1450},
  year={2012},
  publisher={IEEE}
}

@inproceedings{liao2019multiview,
  title={Multiview 2D/3D rigid registration via a point-of-interest network for tracking and triangulation},
  author={Liao, Haofu and Lin, Wei-An and Zhang, Jiarui and Zhang, Jingdan and Luo, Jiebo and Zhou, S Kevin},
  booktitle={In Proceedings of the IEEE/CVF Conference on Computer Vision and Pattern Recognition},
  pages={12638--12647},
  year={2019}
}

@article{wang2017dynamic,
  title={Dynamic 2-D/3-D rigid registration framework using point-to-plane correspondence model},
  author={Wang, Jian and Schaffert, Roman and Borsdorf, Anja and Heigl, Benno and Huang, Xiaolin and Hornegger, Joachim and Maier, Andreas},
  journal={IEEE transactions on medical imaging},
  volume={36},
  number={9},
  pages={1939--1954},
  year={2017},
  publisher={IEEE}
}

@inproceedings{chen2024embedded,
  title={Embedded feature similarity optimization with specific parameter initialization for 2d/3d medical image registration},
  author={Chen, Minheng and Zhang, Zhirun and Gu, Shuheng and Kong, Youyong},
  booktitle={ICASSP 2024-2024 IEEE International Conference on Acoustics, Speech and Signal Processing (ICASSP)},
  pages={1521--1525},
  year={2024},
  organization={IEEE}
}

@article{zhao2024automatic,
  title={Automatic 2D/3D spine registration based on two-step transformer with semantic attention and adaptive multi-dimensional loss function},
  author={Zhao, Huiyu and Zhu, Wangshu and Deng, Xiao and Zhang, Guowang and Zou, Weiwen and others},
  journal={Biomedical Signal Processing and Control},
  volume={95},
  pages={106384},
  year={2024},
  publisher={Elsevier}
}

@inproceedings{grzech2024unsupervised,
  title={Unsupervised Similarity Learning for Image Registration with Energy-Based Models},
  author={Grzech, Daniel and Folgoc, Lo{\"\i}c Le and Azampour, Mohammad Farid and Vlontzos, Athanasios and Glocker, Ben and Navab, Nassir and Schnabel, Julia and Kainz, Bernhard},
  booktitle={International Workshop on Biomedical Image Registration},
  pages={229--240},
  year={2024},
  organization={Springer}
}

@article{chen2024optimization,
  title={An optimization-based baseline for rigid 2d/3d registration applied to spine surgical navigation using cma-es},
  author={Chen, Minheng and Li, Tonglong and Zhang, Zhirun and Kong, Youyong},
  journal={arXiv preprint arXiv:2402.05642},
  year={2024}
}

@inproceedings{chen2025introducing,
  title={Introducing Learning Rate Adaptation CMA-ES into Rigid Intraoperative 2D/3D Registration for Spinal Surgery},
  author={Chen, Minheng and Zhang, Zhirun},
  booktitle={2025 IEEE 22nd International Symposium on Biomedical Imaging (ISBI)},
  pages={1--4},
  year={2025},
  organization={IEEE}
}

@article{toews2017phantomless,
  title={Phantomless auto-calibration and online calibration assessment for a tracked freehand 2-D ultrasound probe},
  author={Toews, Matthew and Wells, William M},
  journal={IEEE transactions on medical imaging},
  volume={37},
  number={1},
  pages={262--272},
  year={2017},
  publisher={IEEE}
}

@inproceedings{zollei20012d,
  title={2D-3D rigid registration of X-ray fluoroscopy and CT images using mutual information and sparsely sampled histogram estimators},
  author={Zollei, L and Grimson, Eric and Norbash, Alexander and Wells, W},
  booktitle={Proceedings of the 2001 IEEE Computer Society Conference on Computer Vision and Pattern Recognition. CVPR 2001},
  volume={2},
  pages={II--II},
  year={2001},
  organization={IEEE}
}

@inproceedings{zhong2020random,
  title={Random erasing data augmentation},
  author={Zhong, Zhun and Zheng, Liang and Kang, Guoliang and Li, Shaozi and Yang, Yi},
  booktitle={Proceedings of the AAAI conference on artificial intelligence},
  volume={34},
  number={07},
  pages={13001--13008},
  year={2020}
}

@inproceedings{grupp2018patch,
  title={Patch-based image similarity for intraoperative 2D/3D pelvis registration during periacetabular osteotomy},
  author={Grupp, Robert B and Armand, Mehran and Taylor, Russell H},
  booktitle={International Workshop on Computer-Assisted and Robotic Endoscopy},
  pages={153--163},
  year={2018},
  organization={Springer}
}

@article{penney1998comparison,
  title={A comparison of similarity measures for use in 2-D-3-D medical image registration},
  author={Penney, Graeme P and Weese, J{\"u}rgen and Little, John A and Desmedt, Paul and Hill, Derek LG and others},
  journal={IEEE transactions on medical imaging},
  volume={17},
  number={4},
  pages={586--595},
  year={1998},
  publisher={IEEE}
}

@inproceedings{knaan2003effective,
  title={Effective intensity-based 2D/3D rigid registration between fluoroscopic X-ray and CT},
  author={Knaan, Dotan and Joskowicz, Leo},
  booktitle={International Conference on Medical Image Computing and Computer-Assisted Intervention},
  pages={351--358},
  year={2003},
  organization={Springer}
}

@article{frysch2021novel,
  title={A novel approach to 2D/3D registration of X-ray images using Grangeat’s relation},
  author={Frysch, Robert and Pfeiffer, Tim and Rose, Georg},
  journal={Medical Image Analysis},
  volume={67},
  pages={101815},
  year={2021},
  publisher={Elsevier}
}

@article{de20163d,
  title={3D--2D image registration for target localization in spine surgery: investigation of similarity metrics providing robustness to content mismatch},
  author={De Silva, T and Uneri, A and Ketcha, MD and Reaungamornrat, S and Kleinszig, G and Vogt, S and Aygun, Nafi and Lo, SF and Wolinsky, JP and Siewerdsen, JH},
  journal={Physics in Medicine \& Biology},
  volume={61},
  number={8},
  pages={3009},
  year={2016},
  publisher={IOP Publishing}
}

@inproceedings{markova2022global,
  title={Global multi-modal 2d/3d registration via local descriptors learning},
  author={Markova, Viktoria and Ronchetti, Matteo and Wein, Wolfgang and Zettinig, Oliver and Prevost, Raphael},
  booktitle={International conference on medical image computing and computer-assisted intervention},
  pages={269--279},
  year={2022},
  organization={Springer}
}

@inproceedings{leroy2023structuregnet,
  title={StructuRegNet: Structure-guided multimodal 2D-3D registration},
  author={Leroy, Amaury and Cafaro, Alexandre and Gessain, Gr{\'e}goire and Champagnac, Anne and Gr{\'e}goire, Vincent and Deutsch, Eric and Lepetit, Vincent and Paragios, Nikos},
  booktitle={International Conference on Medical Image Computing and Computer-Assisted Intervention},
  pages={771--780},
  year={2023},
  organization={Springer}
}

@inproceedings{hou2018computing,
  title={Computing CNN loss and gradients for pose estimation with Riemannian geometry},
  author={Hou, Benjamin and Miolane, Nina and Khanal, Bishesh and Lee, Matthew CH and Alansary, Amir and McDonagh, Steven and Hajnal, Jo V and Rueckert, Daniel and Glocker, Ben and Kainz, Bernhard},
  booktitle={International Conference on Medical Image Computing and Computer-Assisted Intervention},
  pages={756--764},
  year={2018},
  organization={Springer}
}

@article{mi2023sgreg,
  title={SGReg: segmentation guided 3D/2D rigid registration for orthogonal X-ray and CT images in spine surgery navigation},
  author={Mi, Jia and Yin, Wenhao and Zhao, Lei and Chen, Yangfan and Zhou, Yujia and Feng, Qianjin},
  journal={Physics in Medicine \& Biology},
  volume={68},
  number={13},
  pages={135004},
  year={2023},
  publisher={IOP Publishing}
}

@inproceedings{sideri2023mad,
  title={Mad: Modality agnostic distance measure for image registration},
  author={Sideri-Lampretsa, Vasiliki and Zimmer, Veronika A and Qiu, Huaqi and Kaissis, Georgios and Rueckert, Daniel},
  booktitle={International Conference on Medical Image Computing and Computer-Assisted Intervention},
  pages={147--156},
  year={2023},
  organization={Springer}
}

@inproceedings{ronchetti2023disa,
  title={Disa: Differentiable similarity approximation for universal multimodal registration},
  author={Ronchetti, Matteo and Wein, Wolfgang and Navab, Nassir and Zettinig, Oliver and Prevost, Raphael},
  booktitle={International Conference on Medical Image Computing and Computer-Assisted Intervention},
  pages={761--770},
  year={2023},
  organization={Springer}
}

@inproceedings{brandstatter2024rigid,
  title={Rigid Single-Slice-in-Volume registration via rotation-equivariant 2D/3D feature matching},
  author={Brandst{\"a}tter, Stefan and Seeb{\"o}ck, Philipp and F{\"u}rb{\"o}ck, Christoph and Pochepnia, Svitlana and Prosch, Helmut and Langs, Georg},
  booktitle={International Workshop on Biomedical Image Registration},
  pages={280--294},
  year={2024},
  organization={Springer}
}

@inproceedings{mok2024modality,
  title={Modality-agnostic structural image representation learning for deformable multi-modality medical image registration},
  author={Mok, Tony CW and Li, Zi and Bai, Yunhao and Zhang, Jianpeng and Liu, Wei and Zhou, Yan-Jie and Yan, Ke and Jin, Dakai and Shi, Yu and Yin, Xiaoli and others},
  booktitle={Proceedings of the IEEE/CVF Conference on Computer Vision and Pattern Recognition},
  pages={11215--11225},
  year={2024}
}

@inproceedings{zhang2023patient,
  title={A patient-specific self-supervised model for automatic X-Ray/CT registration},
  author={Zhang, Baochang and Faghihroohi, Shahrooz and Azampour, Mohammad Farid and Liu, Shuting and Ghotbi, Reza and Schunkert, Heribert and Navab, Nassir},
  booktitle={International Conference on Medical Image Computing and Computer-Assisted Intervention},
  pages={515--524},
  year={2023},
  organization={Springer}
}

@inproceedings{gopalakrishnan2024intraoperative,
  title={Intraoperative 2d/3d image registration via differentiable x-ray rendering},
  author={Gopalakrishnan, Vivek and Dey, Neel and Golland, Polina},
  booktitle={Proceedings of the IEEE/CVF Conference on Computer Vision and Pattern Recognition},
  pages={11662--11672},
  year={2024}
}

@article{schaffert2020learning,
  title={Learning an attention model for robust 2-D/3-D registration using point-to-plane correspondences},
  author={Schaffert, Roman and Wang, Jian and Fischer, Peter and Borsdorf, Anja and Maier, Andreas},
  journal={IEEE transactions on medical imaging},
  volume={39},
  number={10},
  pages={3159--3174},
  year={2020},
  publisher={IEEE}
}

@inproceedings{jaganathan2023self,
  title={Self-supervised 2D/3D registration for X-ray to CT image fusion},
  author={Jaganathan, Srikrishna and Kukla, Maximilian and Wang, Jian and Shetty, Karthik and Maier, Andreas},
  booktitle={Proceedings of the IEEE/CVF Winter Conference on Applications of Computer Vision},
  pages={2788--2798},
  year={2023}
}

@article{fischler1981random,
  title={Random sample consensus: a paradigm for model fitting with applications to image analysis and automated cartography},
  author={Fischler, Martin A and Bolles, Robert C},
  journal={Communications of the ACM},
  volume={24},
  number={6},
  pages={381--395},
  year={1981},
  publisher={ACM New York, NY, USA}
}

@inproceedings{shrestha2023x,
  title={X-ray to ct rigid registration using scene coordinate regression},
  author={Shrestha, Pragyan and Xie, Chun and Shishido, Hidehiko and Yoshii, Yuichi and Kitahara, Itaru},
  booktitle={International Conference on Medical Image Computing and Computer-Assisted Intervention},
  pages={781--790},
  year={2023},
  organization={Springer}
}

@inproceedings{esteban2019towards,
  title={Towards fully automatic X-ray to CT registration},
  author={Esteban, Javier and Grimm, Matthias and Unberath, Mathias and Zahnd, Guillaume and Navab, Nassir},
  booktitle={International Conference on Medical Image Computing and Computer Assisted Intervention},
  pages={631--639},
  year={2019},
  organization={Springer}
}

@article{grimm2021pose,
  title={Pose-dependent weights and domain randomization for fully automatic X-ray to CT registration},
  author={Grimm, Matthias and Esteban, Javier and Unberath, Mathias and Navab, Nassir},
  journal={IEEE transactions on medical imaging},
  volume={40},
  number={9},
  pages={2221--2232},
  year={2021},
  publisher={IEEE}
}

@INPROCEEDINGS{chen2024fully,
  author={Chen, Minheng and Zhang, Zhirun and Gu, Shuheng and Ge, Zhangyang and Kong, Youyong},
  booktitle={2024 IEEE International Symposium on Biomedical Imaging (ISBI)}, 
  title={Fully Differentiable Correlation-Driven 2D/3D Registration for X-Ray to CT Image Fusion}, 
  year={2024},
  volume={},
  number={},
  pages={1-5},
  doi={10.1109/ISBI56570.2024.10635662}}

@inproceedings{gu2020extended,
  title={Extended capture range of rigid 2D/3D registration by estimating Riemannian pose gradients},
  author={Gu, Wenhao and Gao, Cong and Grupp, Robert and Fotouhi, Javad and Unberath, Mathias},
  booktitle={Machine Learning in Medical Imaging: 11th International Workshop},
  pages={281--291},
  year={2020},
  organization={Springer}
}

@inproceedings{grzech2022variational,
  title={A variational bayesian method for similarity learning in non-rigid image registration},
  author={Grzech, Daniel and Azampour, Mohammad Farid and Glocker, Ben and Schnabel, Julia and Navab, Nassir and Kainz, Bernhard and Le Folgoc, Lo{\"\i}c},
  booktitle={Proceedings of the IEEE/CVF conference on computer vision and pattern recognition},
  pages={119--128},
  year={2022}
}

@inproceedings{qin2019unsupervised,
  title={Unsupervised deformable registration for multi-modal images via disentangled representations},
  author={Qin, Chen and Shi, Bibo and Liao, Rui and Mansi, Tommaso and Rueckert, Daniel and Kamen, Ali},
  booktitle={International Conference on Information Processing in Medical Imaging},
  pages={249--261},
  year={2019},
  organization={Springer}
}

@inproceedings{unberath2018deepdrr,
  title={DeepDRR--a catalyst for machine learning in fluoroscopy-guided procedures},
  author={Unberath, Mathias and Zaech, Jan-Nico and Lee, Sing Chun and Bier, Bastian and Fotouhi, Javad and Armand, Mehran and Navab, Nassir},
  booktitle={International Conference on Medical Image Computing and Computer Assisted Intervention},
  pages={98--106},
  year={2018},
  organization={Springer}
}

@article{momeni2024voxel,
  title={Differentiable Voxel-based X-ray Rendering Improves Sparse-View 3D CBCT Reconstruction},
  author={Momeni, Mohammadhossein and Gopalakrishnan, Vivek and Dey, Neel and Golland, Polina and Frisken, Sarah},
  journal={Machine Learning and the Physical Sciences, NeurIPS 2024},
  booktitle={Machine Learning and the Physical Sciences, NeurIPS 2024},
  year={2024}
}

@article{siddon1985fast,
  title={Fast calculation of the exact radiological path for a three-dimensional CT array},
  author={Siddon, Robert L},
  journal={Medical physics},
  volume={12},
  number={2},
  pages={252--255},
  year={1985},
  publisher={Wiley Online Library}
}

@inproceedings{bui2017x,
  title={X-ray posenet: 6 dof pose estimation for mobile x-ray devices},
  author={Bui, Mai and Albarqouni, Shadi and Schrapp, Michael and Navab, Nassir and Ilic, Slobodan},
  booktitle={2017 IEEE Winter Conference on Applications of Computer Vision (WACV)},
  pages={1036--1044},
  year={2017},
  organization={IEEE}
}

@book{murray2017mathematical,
  title={A mathematical introduction to robotic manipulation},
  author={Murray, Richard M and Li, Zexiang and Sastry, S Shankar},
  year={2017},
  publisher={CRC press}
}

@article{billot2024se,
  title={SE (3)-equivariant and noise-invariant 3D rigid motion tracking in brain MRI},
  author={Billot, Benjamin and Dey, Neel and Moyer, Daniel and Hoffmann, Malte and Turk, Esra Abaci and Gagoski, Borjan and Grant, P Ellen and Golland, Polina},
  journal={IEEE transactions on medical imaging},
  year={2024},
  publisher={IEEE}
}

@article{wilson2014spherical,
  title={Spherical and hyperbolic embeddings of data},
  author={Wilson, Richard C and Hancock, Edwin R and Pekalska, El{\.z}bieta and Duin, Robert PW},
  journal={IEEE transactions on pattern analysis and machine intelligence},
  volume={36},
  number={11},
  pages={2255--2269},
  year={2014},
  publisher={IEEE}
}

@inproceedings{zhao2023spherical,
  title={Spherical space feature decomposition for guided depth map super-resolution},
  author={Zhao, Zixiang and Zhang, Jiangshe and Gu, Xiang and Tan, Chengli and Xu, Shuang and Zhang, Yulun and Timofte, Radu and Van Gool, Luc},
  booktitle={Proceedings of the IEEE/CVF International Conference on Computer Vision},
  pages={12547--12558},
  year={2023}
}

@article{meng2025parametric,
  title={Parametric Bi-invariant Learning for Improved Precision in 2D/3D Image Registration},
  author={Meng, Ruoyu and Chen, Chunxiao and Lu, Ming and Fu, Xue and Xiao, Yueyue and Wang, Kunpeng and Zou, Yuan and Li, Yang},
  journal={Biomedical Signal Processing and Control},
  volume={105},
  pages={107603},
  year={2025},
  publisher={Elsevier}
}

@article{thomas2014approaching,
  title={Approaching dual quaternions from matrix algebra},
  author={Thomas, Federico},
  journal={IEEE Transactions on Robotics},
  volume={30},
  number={5},
  pages={1037--1048},
  year={2014},
  publisher={IEEE}
}

@inproceedings{mahendran20173d,
  title={3d pose regression using convolutional neural networks},
  author={Mahendran, Siddharth and Ali, Haider and Vidal, Ren{\'e}},
  booktitle={Proceedings of the IEEE international conference on computer vision workshops},
  pages={2174--2182},
  year={2017}
}

@article{zhang2024spineclue,
title = {SpineCLUE: Automatic vertebrae identification using contrastive learning and uncertainty estimation},
journal = {Artificial Intelligence in Medicine},
volume = {171},
pages = {103285},
year = {2026},
issn = {0933-3657},
doi = {https://doi.org/10.1016/j.artmed.2025.103285},
url = {https://www.sciencedirect.com/science/article/pii/S0933365725002209},
author = {Sheng Zhang and Hongxuan Li and Minheng Chen and Mingying Li and Miao Liu and Junxian Wu and Cheng Xue and Youyong Kong}
}

@article{gopalakrishnan2025rapid,
  title={Rapid patient-specific neural networks for intraoperative X-ray to volume registration},
  author={Gopalakrishnan, Vivek and Dey, Neel and Chlorogiannis, David-Dimitris and Abumoussa, Andrew and Larson, Anna M and Orbach, Darren B and Frisken, Sarah and Golland, Polina},
  journal={ArXiv},
  pages={arXiv--2503},
  year={2025}
}

@inproceedings{jaganathan2021deep,
  title={Deep iterative 2d/3d registration},
  author={Jaganathan, Srikrishna and Wang, Jian and Borsdorf, Anja and Shetty, Karthik and Maier, Andreas},
  booktitle={International Conference on Medical Image Computing and Computer-Assisted Intervention},
  pages={383--392},
  year={2021},
  organization={Springer}
}

@article{salehi2018real,
  title={Real-time deep pose estimation with geodesic loss for image-to-template rigid registration},
  author={Salehi, Seyed Sadegh Mohseni and Khan, Shadab and Erdogmus, Deniz and Gholipour, Ali},
  journal={IEEE transactions on medical imaging},
  volume={38},
  number={2},
  pages={470--481},
  year={2018},
  publisher={IEEE}
}

@article{grupp2020automatic,
  title={Automatic annotation of hip anatomy in fluoroscopy for robust and efficient 2D/3D registration},
  author={Grupp, Robert B and Unberath, Mathias and Gao, Cong and Hegeman, Rachel A and Murphy, Ryan J and Alexander, Clayton P and Otake, Yoshito and McArthur, Benjamin A and Armand, Mehran and Taylor, Russell H},
  journal={International journal of computer assisted radiology and surgery},
  volume={15},
  pages={759--769},
  year={2020},
  publisher={Springer}
}

@article{pernus20133d,
  title={3D-2D registration of cerebral angiograms: A method and evaluation on clinical images},
  author={Pernus, Franjo and others},
  journal={IEEE transactions on medical imaging},
  volume={32},
  number={8},
  pages={1550--1563},
  year={2013},
  publisher={IEEE}
}

@article{deng2021ctspine1k,
  title={Ctspine1k: A large-scale dataset for spinal vertebrae segmentation in computed tomography},
  author={Deng, Yang and Wang, Ce and Hui, Yuan and Li, Qian and Li, Jun and Luo, Shiwei and Sun, Mengke and Quan, Quan and Yang, Shuxin and Hao, You and others},
  journal={arXiv preprint arXiv:2105.14711},
  year={2021}
}

@article{campbell2015automated,
  title={An automated method for landmark identification and finite-element modeling of the lumbar spine},
  author={Campbell, Julius Quinn and Petrella, Anthony J},
  journal={IEEE Transactions on Biomedical Engineering},
  volume={62},
  number={11},
  pages={2709--2716},
  year={2015},
  publisher={IEEE}
}

@article{powell2009bobyqa,
  title={The BOBYQA algorithm for bound constrained optimization without derivatives},
  author={Powell, Michael JD and others},
  journal={Cambridge NA Report NA2009/06, University of Cambridge, Cambridge},
  volume={26},
  pages={26--46},
  year={2009}
}

@inproceedings{moyer2021equivariant,
  title={Equivariant filters for efficient tracking in 3d imaging},
  author={Moyer, Daniel and Abaci Turk, Esra and Grant, P Ellen and Wells, William M and Golland, Polina},
  booktitle={Medical Image Computing and Computer Assisted Intervention--MICCAI 2021: 24th International Conference, Strasbourg, France, September 27--October 1, 2021, Proceedings, Part IV 24},
  pages={193--202},
  year={2021},
  organization={Springer}
}

@article{li2025automatic,
  title={Automatic x-ray to CT registration using embedding reconstruction and lite cross-attention},
  author={Li, Tonglong and Chen, Minheng and Li, Mingying and Li, Chuanyou and Kong, Youyong},
  journal={Medical Physics},
  year={2025},
  publisher={Wiley Online Library}
}

@article{downs2025improving,
  title={Improving automatic cerebral 3D-2D CTA-DSA registration},
  author={Downs, Charles and Sluijs, P and Cornelissen, Sandra AP and Nijenhuis, Frank te and Zwam, Wim H van and Gopalakrishnan, Vivek and Zhang, Xucong and Su, Ruisheng and Walsum, Theo van},
  journal={International Journal of Computer Assisted Radiology and Surgery},
  pages={1--10},
  year={2025},
  publisher={Springer}
}

@inproceedings{gopalakrishnan2022fast,
  title={Fast auto-differentiable digitally reconstructed radiographs for solving inverse problems in intraoperative imaging},
  author={Gopalakrishnan, Vivek and Golland, Polina},
  booktitle={Workshop on Clinical Image-Based Procedures},
  pages={1--11},
  year={2022},
  organization={Springer}
}

@article{miao2016cnn,
  title={A CNN regression approach for real-time 2D/3D registration},
  author={Miao, Shun and Wang, Z Jane and Liao, Rui},
  journal={IEEE transactions on medical imaging},
  volume={35},
  number={5},
  pages={1352--1363},
  year={2016},
  publisher={IEEE}
}

@inproceedings{shrestha2024rayemb,
  title={RayEmb: Arbitrary Landmark Detection in X-Ray Images Using Ray Embedding Subspace},
  author={Shrestha, Pragyan and Xie, Chun and Yoshii, Yuichi and Kitahara, Itaru},
  booktitle={Proceedings of the Asian Conference on Computer Vision},
  pages={665--681},
  year={2024}
}
}
\clearpage
\setcounter{page}{1}
\maketitlesupplementary

% \noindent\textbf{Overview.}
\noindent In this {supplementary material}, we first provide the preprocessing details of the X-ray data used in the experiments in \cref{sec::preprocess}. 
Following that, we include the detailed descriptions of three datasets in \cref{sec::dataset}.
\cref{sec::imple} describes the implementation details of the proposed method.
Finally, in~\cref{sec::random},we provide the details of domain randomization strategy we used in the this work.

\section{Preprocessing of Real X-ray Data}
\label{sec::preprocess}
As mentioned in numerous prior studies~\cite{gopalakrishnan2024intraoperative,unberath2018deepdrr,momeni2024voxel}, X-ray imaging captures the intensity attenuation of X-rays as they traverse a medium, whereas DRRs quantify the absorption of X-ray energy by the medium.
Therefore, in DRRs, higher-density bone tissue appears brighter, while lower-density soft tissue appears relatively darker. This contrast is inverted in X-ray images, where bone structures appear darker due to greater X-ray absorption, while soft tissue appears brighter due to lower attenuation.
To ensure that the grayscale distribution of the DRR generated using Siddon's method~\cite{siddon1985fast} aligns with that of X-ray images governed by the Beer-Lambert law, we applied the following inversion strategy:
\begin{equation}
    \Tilde{I} = 1 - \frac{\log(1+I)}{\log(1+I_0)}
\end{equation}
where $I_0$ represents the initial energy of the radiation beam, and all X-ray image pixel values are normalized to their maximum intensity. The transformation $\log(1+I)$ ensures numerical stability by preventing negative logarithmic values while enhancing visual contrast. Additionally, it promotes a Gaussian-like intensity distribution, improving the consistency of the processed images.

\section{Datasets}
\label{sec::dataset}
We perform experiments on three publicly available datasets to evaluate the effectiveness of the proposed method:
\begin{itemize}
\item  \textbf{DeepFluoro}: The DeepFluoro dataset contains six pelvic CT scans with a total of 366 real fluoroscopic X-ray images~\cite{grupp2020automatic}. It provides the calibrated intrinsic matrix of the C-arm imaging system along with the ground truth extrinsic matrix for each X-ray image. Additionally, 14 manually annotated landmarks are available for each CT scan.
\item \textbf{Ljubljana}:   A clinical dataset~\cite{pernus20133d} contains ten clinical 3D Cone-Beam Computed Tomography (CBCT) subtracted angiography scans of patients undergoing cerebral endovascular treatments. Each CBCT scan is provided with two corresponding X-ray images. Calibrated intrinsic and extrinsic matrices are provided.
\item \textbf{CTSpine1k}: The CTSpine1K dataset~\cite{deng2021ctspine1k} consists of 1005 CT volumes with segmentation masks collected from four different open sources, capturing diverse appearance variations. After quality control, 651 CT scans were retained. Following segmentation and denoising preprocessing, only the lumbar spine was extracted. 
The contour points of the spine surface were obtained by segmentation using Hounsfield unit (HU) threshold, morphological refinement, and 3D surface extraction using the Marching Cubes algorithm.
The automatic landmark identification follows a strategy similar to~\cite{campbell2015automated,zhang2024spineclue}.
For each CT scan, 500 DRR images with random poses were generated using~\cite{unberath2018deepdrr}. The intrinsic matrix used in the simulation environment is calibrated to match the settings of a mobile C-arm imaging device.
\end{itemize}
\section{Implementation Details}
\label{sec::imple}
For patient-specific 2D/3D registration, we selected ResNet-18 as the backbone for the pose regressor $\mathcal{R}(\cdot)$, using the Adam optimizer with a learning rate of 1e-3 and a cosine scheduler. The spherical similarity network, used for subsequent fine-tuning, was trained with the AdamW optimizer, also with a learning rate of 1e-3, but utilizing a cyclic scheduler. The pose regressor and similarity network were trained in a self-supervised manner using 500k and 200k synthetic images generated in real-time, respectively.
For the patient-agnostic scenario, the parameter settings remained largely the same, except that the pose regressor was replaced with RTPIv3~\cite{chen2024embedded}, which incorporates spatial information from CT. Additionally, the number of self-supervised training iterations was increased to 750k for the pose regressor and 300k for the similarity network, to better generalize across different subjects.
The termination criterion for the differentiable LM optimization is set such that the standard deviation of $\Delta\theta_i$ over the last ten iterations is less than 1e-2. 
This threshold was determined through grid search across all datasets to ensure an optimal balance between convergence stability and computational efficiency.
Our experiments were conducted on a PC equipped with an NVIDIA RTX 3090 GPU and an Intel Core i7 CPU. Due to memory limitations, the batch sizes for training the pose regressor and the spherical similarity network were set to 4 and 1, respectively.
\section{Domain Randomization Strategy}
\label{sec::random}
Similar to~\cite{grimm2021pose,gopalakrishnan2024intraoperative,zhang2023patient}, to alleviate the domain gap between synthetic image and real X-ray image, domain randomization strategy is adopted during training:

\begin{itemize}
    \item \textbf{Image Smoothing}: Perform random smoothing on the image with a kernel size of 3×3 or 5×5, selected with a probability of 50\%.
    \item \textbf{Noise Injection}: Inject Gaussian noise into the image with a mean sampled uniformly from $[-0.15 \cdot \text{max}, 0.1 \cdot \text{max}]$.
    \item \textbf{Normalization}: Apply lower and upper bound normalization with intervals sampled as $[-0.04 \cdot \text{max}, 0.02 \cdot \text{max}]$ and $[0.9 \cdot \text{max}, 1.05 \cdot \text{max}]$, respectively.
    \item \textbf{Linear Scaling}: Scale the intensity linearly, with the scaling factor sampled uniformly from $[0.9, 1.05]$.
    \item \textbf{Gamma Adjustment}: Perform gamma correction, with the $\gamma$ value sampled uniformly from $[0.7, 1.3]$.
    \item \textbf{Nonlinear Scaling}: Scale the image nonlinearly using the function $a \cdot \sin(b \cdot x + c)$, where $a$ and $b$ are sampled uniformly from the range $[0.8, 1.1]$, and $c$ is sampled uniformly from $[-0.5, 0.4]$.
    \item \textbf{Random Erasing}~\cite{zhong2020random}: Randomly erases a rectangular region of the image with an area uniformly sampled from $[0.02 \cdot \text{area}, 0.4 \cdot \text{area}]$, and an aspect ratio sampled uniformly from $[0.3, 1]$, filling the region with the mean intensity of the whole image.
\end{itemize}

\end{document}